\pdfoutput=1
\documentclass[10pt,twocolumn,letterpaper]{article}

\usepackage{iccv}
\usepackage{times}
\usepackage{epsfig}
\usepackage{graphicx}
\usepackage{amsmath}
\usepackage{amssymb}

\usepackage{graphicx}
\usepackage{amsmath}
\usepackage{amsfonts}
\usepackage{dsfont}
\usepackage{amssymb}
\usepackage{booktabs}
\usepackage{caption}

\usepackage{color}
\usepackage{xcolor}
\usepackage{xspace}

\newcommand{\Hquad}{\hspace{0.5em}}

\usepackage{bm}
\usepackage{multirow}

\usepackage[breaklinks=true,bookmarks=false]{hyperref}

\iccvfinalcopy 

\def\iccvPaperID{1310} 

\ificcvfinal\fi

\begin{document}

\title{Ada3D : Exploiting the Spatial Redundancy with \\Adaptive Inference for Efficient 3D Object Detection}

\author{Tianchen Zhao$^{12}$, Xuefei Ning$^{1*}$, Ke Hong$^{1}$, Zhongyuan Qiu$^{2}$, Pu Lu$^{1}$, Yali Zhao$^{2}$, \\
Linfeng Zhang$^{1}$, Lipu Zhou$^{3}$, Guohao Dai$^{4}$, Huazhong Yang$^{1}$, Yu Wang$^{1}\thanks{Corresponding Authors}$\\
$^1$Tsinghua University, $^2$Novauto, $^3$Meituan, $^4$Shanghai Jiao Tong University\\
\tt\small \{suozhang1998,foxdoraame\}@gmail.com \quad \{zhongyuan.qiu,yali.zhao\}@novauto.com.cn  \\
\tt\small \{zhanglinfeng1997,zhoulipu\}@outlook.com \quad daiguohao@sjtu.edu.cn \\  
\tt\small \{hongk21,lup22,yanghz,yu-wang\}@mail.tsinghua.edu.cn
}

\maketitle
\ificcvfinal\thispagestyle{empty}\fi

\begin{abstract}
Voxel-based methods have achieved state-of-the-art performance for 3D object detection in autonomous driving. However, their significant computational and memory costs pose a challenge for their application to resource-constrained vehicles. One reason for this high resource consumption is the presence of a large number of redundant background points in Lidar point clouds, resulting in spatial redundancy in both 3D voxel and BEV map representations. To address this issue, we propose an adaptive inference framework called \textbf{Ada3D}, which focuses on reducing the \textbf{spatial redundancy} to compress the model's computational and memory cost. Ada3D adaptively filters the redundant input, guided by a lightweight importance predictor and the unique properties of the Lidar point cloud. Additionally, we maintain the BEV features' intrinsic sparsity by introducing the Sparsity Preserving Batch Normalization. With Ada3D, we achieve $\mathbf{40\%}$ reduction for 3D voxels and decrease the density of 2D BEV feature maps from 100\% to $\mathbf{20\%}$ without sacrificing accuracy. Ada3D reduces the model computational and memory cost by $\mathbf{5\times}$, and achieves $\mathbf{1.52\times}$ / $\mathbf{1.45\times}$ end-to-end GPU latency and $\mathbf{1.5\times}$ / $\mathbf{4.5\times}$ GPU peak memory optimization for the 3D and 2D backbone respectively.
\end{abstract}

\begin{figure}[th]
    \centering
    \includegraphics[width=\linewidth]{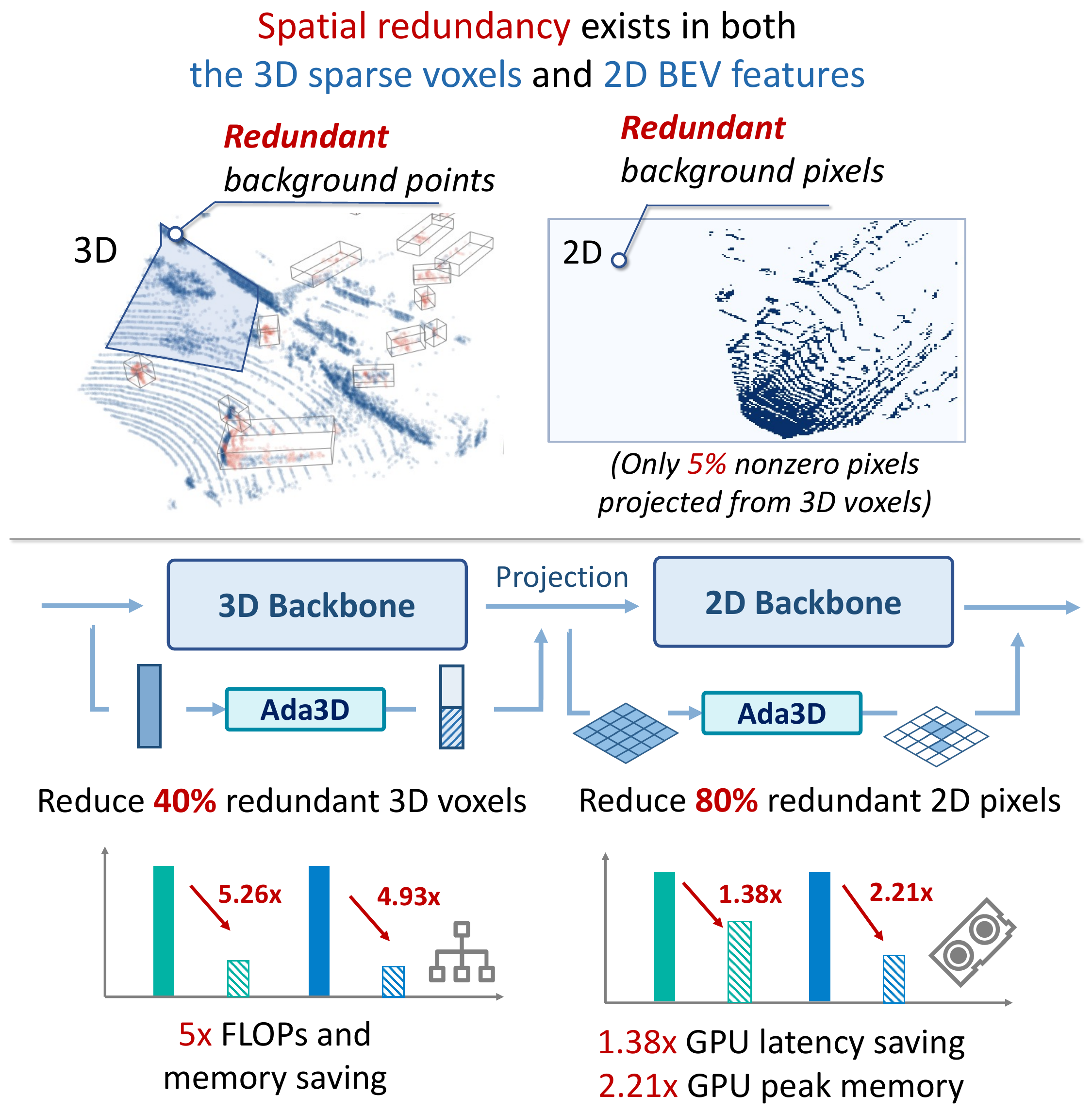}
    \caption{Ada3D is an adaptive inference framework that exploits the spatial redundancy for both the 3D voxel and 2D BEV features. 
    }
    \label{fig:teaser}
\end{figure}

\section{Introduction}
\label{sec:intro}

The perception of the 3D scene plays a vital role in autonomous driving systems. It's essential that the perception of the surrounding 3D scene is both quick and accurate, which places high demands on both performance and latency for perception methods. 

Voxel-based 3D deep learning methods convert the input point cloud into sparse voxels by quantizing them into regular grids, and achieve state-of-the-art performance~\cite{BEV-survey}. However, current voxel-based methods struggle to meet the real-time demand on self-driving cars due to constrained resources~\cite{pointacc}. As a result, it is crucial to improve the efficiency of voxel-based 3D perception methods (e.g., reduce the GPU latency and peak memory).


There are two main factors contributing to the excessively long processing time for 3D perception methods. Firstly, the model size is excessive, and it contains time-consuming operations such as 3D sparse convolution~\cite{pointacc}. Secondly, the algorithm needs to process a large amount of input points (e.g., 30K for nuScenes). Prior researches focus on solving the former issue by compressing the model both at the operation-level~\cite{not_all_neighbor,sps-conv} and architecture-level~\cite{spvnas,pointdistiller}. In this paper, we take a different approach and improve the model's efficiency from the data level.

The typical pipeline of voxel-based 3D detector is displayed in Fig.~\ref{fig:teaser}, the 3D backbone extracts feature from the input point cloud. The 3D features are then projected to bird-eye-view (BEV) space along the z-axis and further processed by the 2D backbone with normal 2D convolutions. 

\begin{figure}[t]
    \centering
    \includegraphics[width=0.95\linewidth]{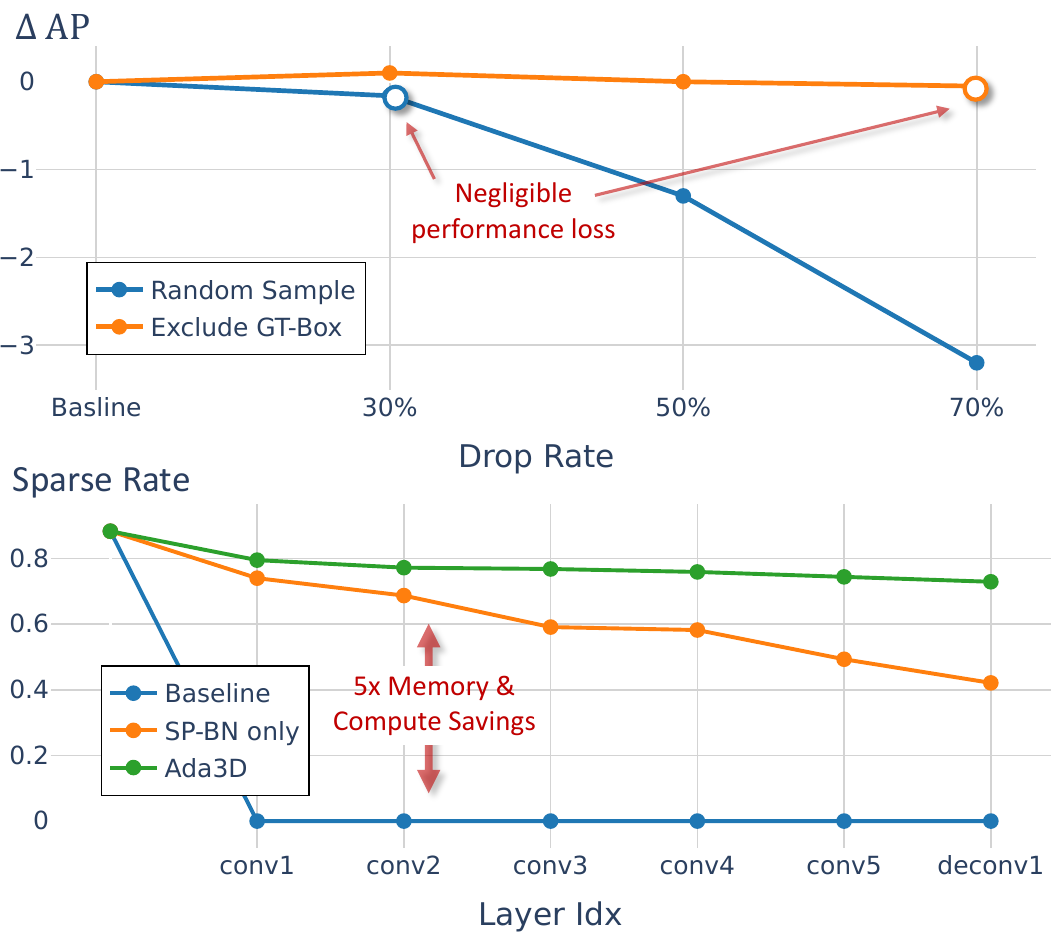}
    \caption{\textbf{Empirical evidence of spatial redundancy in 3D and 2D data.} Upper: The KITTI Cars Moderate AP under different drop rates with random dropping and ground-truth excluded dropping. Lower: The sparsity of different layer's 2D BEV features.}
    \label{fig:oracle}
\end{figure}

We discover that there exists spatial redundancy for both the 3D voxel and 2D BEV features. For 3D voxels, As shown in Fig.~\ref{fig:teaser}, a large number of points in the input point cloud represents the road plane and buildings, which are redundant ``background'' for 3D detection. We further validate the redundancy of the point cloud with quantitative results in Fig.~\ref{fig:oracle}. When we randomly drop 30\% of the input points or 70\% of the points excluding those within the ground-truth bounding box (the ``foreground''), we only observe a subtle drop in performance. Existing 3D CNNs treat all input points equally, thus wasting a substantial amount of computation and memory on the less-informative background area.  
Regarding 2D BEV features, as shown in Fig.~\ref{fig:teaser}, only a small portion of (e.g., 5\% for KITTI) pixels have projected feature values in the BEV space, while others are background pixels with zero value. However, current methods treat these sparse BEV features as dense and apply normal CNN to them. As can be observed in the lower part of Fig.~\ref{fig:oracle}, the feature map loses sparsity after the first batch normalization layer, which fails to utilize the sparse nature of the Lidar-projected BEV feature map. 

To compress the data's spatial redundancy, we propose an adaptive inference method \textbf{Ada3D}. We adopt adaptive inference to both the 3D and 2D backbone and selectively filter out redundant 3D voxels and 2D BEV features during inference. We employ a lightweight predictor to evaluate the importance of input features in the BEV space. The predictor score is combined with the density of the Lidar point cloud to determine which features to drop. In addition, we introduce a simple yet effective technique called sparsity-preserving batch normalization, which efficiently eliminates background pixels and preserves sparsity for 2D BEV features.
Through adaptively skipping redundant features, Ada3D reduces the computational and memory costs of the model by 5$\times$ and achieves 1.4 $\times$ end-to-end speedup and 2.2$\times$ GPU peak memory optimization on RTX3090 without compromising performance.

The contributions of this paper could be summarized into three aspects, as follows:
\begin{enumerate}
    \item We introduce the adaptive inference method Ada3D that leverages spatial redundancy for efficient 3D object detection.
    \item We design a shared predictor to evaluate the importance of input features, and combine the predictor score with point cloud density as the criterion for dropping redundant features. 
    \item We propose sparsity-preserving batch normalization to maintain the sparsity for the 2D backbone. 
\end{enumerate}



\section{Related Works}
\label{sec:rw}

\begin{figure*}[h]
    \centering
    \includegraphics[width=0.95\textwidth]{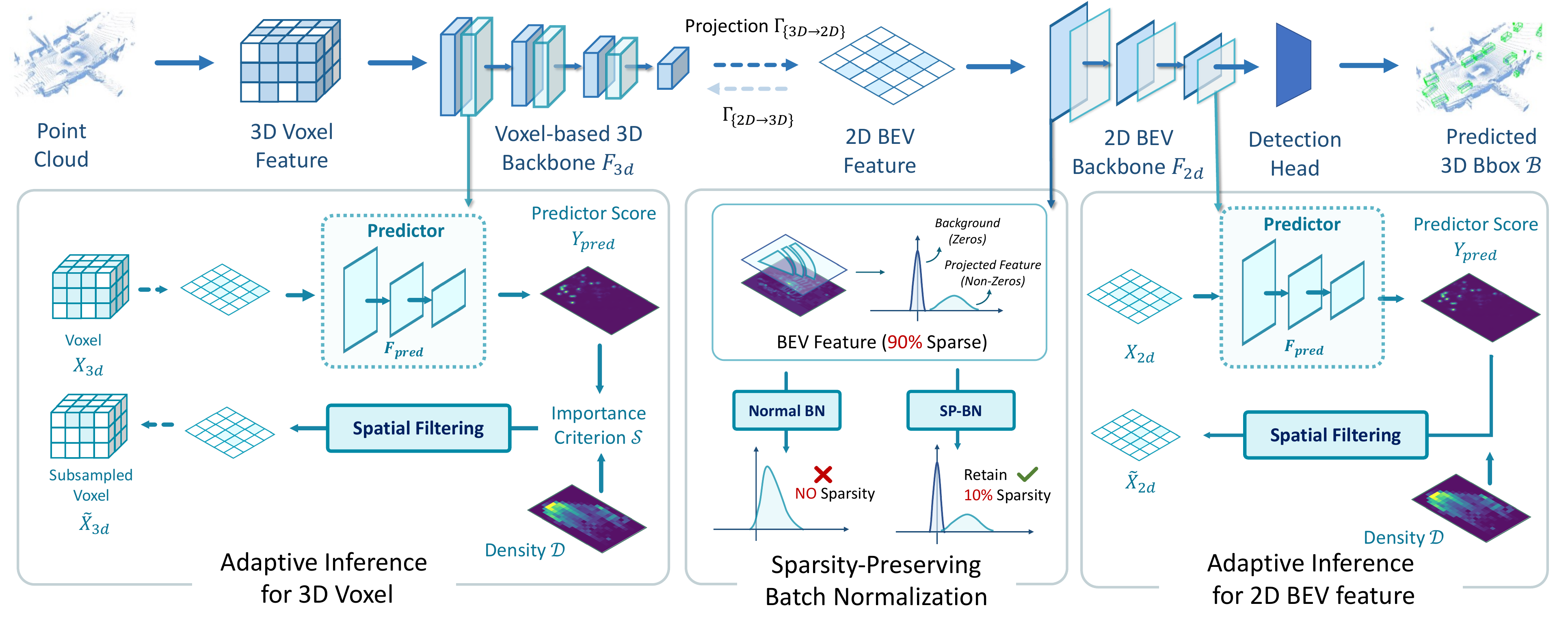}
    \caption{\textbf{The overall framework of Ada3D.} Adaptive inference is conducted in both the 3D and 2D backbone. The spatial filtering module combines the predictor score and 3D point cloud's density to drop the redundant parts. Furthermore, the SP-BN is introduced to omit the background pixels in 2D backbone and retain sparsity.}
    \label{fig:flow}
\end{figure*}

\subsection{Voxel-based 3D Detection Methods}

Voxel-based methods convert the point cloud into regular grids. SECOND~\cite{second} utilizes the 3D sparse convolution for feature extraction. CenterPoint~\cite{centerpoint} is a  single-stage detector that leverages a keypoint detector to detect box centers. PV-RCNN~\cite{pvrcnn} combines the point and voxel features and utilizes a two-staged framework for precise detection. While the voxel-based detectors achieve state-of-the-art results, their high computational and memory costs impede their application on self-driving cars. Ada3D aims to alleviate this issue through adaptive inference. 



\subsection{Adaptive inference for 2D image}

In the field of 2D perception, adaptive inference methods reduce spatial redundancy for 2D images. Figurnov \emph{et. al.}~\cite{figurnov2017spatially} dynamically adjust depth for different regions. A series of methods~\cite{channel_gating,spatial_adaptive,spatial-aware-dynamic} learn to adaptively skip redundant channel/pixels. GFNet~\cite{GFNet} employs reinforcement learning to locate the discriminant regions. Ada3D applies adaptive inference to the 3D perception, and adaptively filters redundant 3D voxels and BEV features.

\subsection{Efficient 3D Detection Methods}

Some prior studies aim to enhance the efficiency of 3D detectors. 
SPVNAS~\cite{spvnas} employs neural architecture search to search to find suitable depth and width for the 3D model. Lee \emph{et. al.}~\cite{not_all_neighbor} propose a point-distribution pruning method on 3D convolution kernel. SPS-Conv~\cite{sps-conv} prunes the output mapping for sparse convolution based on the feature magnitude. RSN~\cite{RSN} designs network module to prune unnecessary part in range view lidar image. A series of fully sparse detectors: FSD~\cite{FSD}, FSD++~\cite{FSD++}, VoxelNeXT~\cite{voxelnext} design novel architectures that eliminate the dense BEV backbone. 
These methods optimize the efficiency of 3D detectors from the perspective of compressing model redundancy. Differently, Ada3D focuses on reducing spatial redundancy and could work on par with these methods.  


\section{Methods}
\label{sec:method}


\subsection{Voxel-based Detection with Adaptive Inference}
\label{sec:method-detection-flow}

Figure~\ref{fig:flow} illustrates the overall framework of Ada3D. The 3D object detection task aims to predict 3D bounding boxes $\mathcal{B}=\{b_k\}$
from the point cloud $\mathcal{P}=\{(x,y,z,r)_i\}$. 
The voxel-based 3D detectors~\cite{pvrcnn,second,centerpoint} quantize the point cloud into regular grids. Without loss of generality, we omit the batch dimension in the following equations. The voxelization generates sparse voxels $\mathbf{X_{3d}} \in \mathbb{R}^{N \times C}$ of voxel numbers $N$ and feature channels $C$. 
The 3D voxel backbone $\mathcal{F}_{3d}$ applies 3D sparse convolution~\cite{sparseconvnet} on the voxels to extract point cloud feature. We use the $\mathbf{X}_{i,c}$ to represent the $c$-th channel of $i$-th voxel feature, and the $c'$ channel of the $j$-th output voxel can be described as:

\begin{equation}
  \begin{split}
  \mathbf{Y}_{j,c'} = \sum_{k}\sum_{c} W_{k,c,c'} X_{R_{k,j},k,c}, \\
  \end{split}
  \label{equ:3d_sparse_conv}
\end{equation}
where $R_{k,j}$ is the input index $i$ given the output index $j$ and kernel offset $k$, $W_{k,c,c'}$ denotes the kernel offset $k$'s weight. 

The processed 3D feature $\mathbf{\tilde{X}}_{\text{3D}}$ is then projected to the BEV plane through sum pooling along the z-axis to generate 2D features $\mathbf{X}_{\text{2D}}\in \mathbb{R}^{C \times W \times H}$.  We define $\Gamma_{3D\to2D}$ as the mapping from 3D voxels to 2D BEV pixels, and $\Gamma_{2D\to3D}$ describes the invert mapping.
The 2D BEV backbone $\mathbf{F}_{\text{2D}}$ is applied to further extract the 2D BEV feature. Finally, the detection head $\mathbf{F}_{\text{head}}$ predicts the 3D bounding box. 

The adaptive inference is adopted in both the 3D and 2D backbone. For simplicity, we omit the channel dimension $C$ for feature $\mathbf{X}$ for the equations below, since all channels share the same spatial filtering pattern. We describe the layer indexes where the predictor is applied with the layer index $\mathcal{I}_{3D}=\{l_{3D}^{(1)}, ..., l_{3D}^{(n)}\}$ and $\mathcal{I}_{2D}=\{l_{2D}^{(1)}, ..., l_{2D}^{(n)}\}$. 
The adaptive inference for 3D backbone could be described as:
\begin{equation}
  \begin{split}
  \mbox{for} \Hquad l_{3D}^{(i)} \in \mathcal{I}_{3D}: \mathbf{X}^{(l_{3D}^{(i)})}_{3D}& = \mathbf{F}_{3D}^{(l_{3D}^{(i)})} (\mathbf{\tilde{X}}^{(l_{3D}^{(i)}-1)}_{3D}), \Hquad \mbox{where} \\
  \mathbf{\tilde{X}}^{(l_{3D}^{(i)}-1)}_{3D} = \Gamma_{2D\to3D}(\mathbf{F}_{\text{drop}}&(\Gamma_{3D\to2D}(\mathbf{X}^{(l_{3D}^{(i)}-1)}_{3D}), \mathbf{S})) \odot \mathbf{X}^{(l_{3D}^{(i)}-1)}_{3D}, \\
  \mathbf{S} = \mathbf{F}_{\text{score}}&(\Gamma_{3D\to2D}(\mathbf{X}^{(l_{3D}^{(i)}-1)}_{3D})),\\
  \end{split}
  \label{equ:adaptive_inference}
\end{equation}
where the $\mathbf{S} \in \mathbb{R}^{W \times H}$ represents the importance score for BEV pixels, which is generated by $\mathbf{F}_{\text{score}}$ that combines the predictor output and 3D point cloud's density. The $\mathbf{F}_{\text{score}}$ takes the 2D BEV input projected from the input 3D voxel feature $\mathbf{X}^{(l_{3D}^{(i)}-1)}_{3D} \in \mathbb{R}^N$. Given the drop ratio $R_{\text{drop}}$, the spatial filtering process $\mathbf{F}_{\text{drop}}$ drops the most redundant portion of features in the BEV space based on the importance score $\mathbf{S}$. It generates the one-hot mask that indicates whether the given location should be kept or discarded. The mask is then broadcasted back to the voxel space through $\Gamma_{2D\to3D}$ and element-wisely multiplied with the original 3D voxel feature to generate subsampled 3D voxel feature $\mathbf{\tilde{X}}^{(l_{3D}^{(i)}-1)}_{3D}$. Note that the equation describes the algorithmic simulation of spatial filtering, while in the actual GPU processing, features with zero values in $\mathbf{\tilde{X}}^{(l_{3D}^{(i)}-1)}_{3D}$ are excluded to achieve actual hardware acceleration, i.e., their computation and storage are skipped. More details about $\mathbf{F}_{\text{drop}}$ and $\mathbf{F}_{\text{score}}$ will be discussed in Sec.~\ref{sec:method-IP} and Sec.~\ref{sec:method-DG}. 

Similarly, the adaptive inference for the 2D BEV backbone is applied at $\mathcal{I}_{2D}=\{l_{2D}^{(1)}, ..., l_{2D}^{(n)}\}$ layers with similar process described in Equ.~\ref{equ:adaptive_inference} without the transformation $\Gamma_{2D\to3D}, \Gamma_{3D\to2D}$ between the voxel and the BEV space.


\subsection{Importance Predictor Design}
\label{sec:method-IP}

As discussed in Equ.~\ref{equ:adaptive_inference} in Sec.~\ref{sec:method-detection-flow}, the $\mathbf{F}_{score}$ is used for evaluating the input feature to identify its redundant parts. In Ada3D, we adopt a lightweight CNN as the spatial-wise importance predictor in BEV space to predict pixel-wise importance score from the input feature.

\textbf{Inference.} The predictor inference for 3D voxel feature is described as:
\begin{equation}
  \begin{split}
  \mathbf{Y}_{\text{pred}} = \mathbf{F}_{\text{pred}}((\mathbf{X}_{\text{BEV}});\Theta_{\text{pred}}), \\
  \end{split}
  \label{equ:predictor_infer}
\end{equation}
where $\mathbf{F}_{\text{pred}}$ is the predictor with the parameter $\Theta_{\text{pred}}$. The predictor's output is a single channel heatmap $\mathbf{Y}_{\text{pred}} \in \left[0,1\right]^{W\times H}$. 
We choose to design the predictor in the BEV space instead of 3D space, as the perception is mainly conducted in the former. Intuitively, there exists less redundancy in the vertical space, and the efficiency improvement of compressing it is restricted. Also, estimating the importance of the whole 3D space is more challenging. 
In order to effectively and efficiently evaluate the importance, we design a lightweight predictor that is shared for different layers at both the 3D and 2D backbone. It consists of multiple group convolutions ~\cite{shufflenet} with reduced parameters and computational complexity. Besides, the resolution of the predictor is selected as 1/8 of the original original BEV resolution. The computaional cost of the predictor's is less than 1\% of the 2D backbone, thereby bringing negilible overhead. 

\textbf{Training.} 
Our oracle experiment in Fig.~\ref{fig:oracle} shows that the performance only decreases slightly when dropping a notable amount of points outside the ground-truth bounding boxes. It reveals that the center of the bounding box is of high importance and the importance spreads to the local region. 
Therefore, following CenterPoint~\cite{centerpoint}, we generate the ground-truth heatmap $M_{\text{gt}}$ for the predictor by rendering a 2D Gaussian circle with a peak located at each bounding box center $(u,v)$, which could be formulated as follows:
\begin{equation}
  \begin{split}
  \mathbf{M}_{\text{gt}}=\sum_{b_i} \mathcal{G}((u,v)_i,\sigma), \\
  \end{split}
  \label{equ:predictor_train}
\end{equation}
where $b_i$ is the ground-truth bounding box, and $\mathcal{G}$ is the 2D gaussian function with radius $\sigma$. 
The mean squared error (MSE) loss is adopted for predictor training.

\subsection{Density-guided Spatial Filtering}
\label{sec:method-DG}

The spatial filtering $\mathbf{F}_{drop}$ in Equ.~\ref{equ:adaptive_inference} in Sec.~\ref{sec:method-detection-flow} describes the process of dropping the most redundant $R_{\text{drop}}$ of the input features based on the importance criterion $\mathbf{S}$. We combine the predictor score with the point cloud density to determine where to drop. 

The predictor score $\mathbf{Y}_{pred}$ could effectively represent the relative importance of the input feature. However, due to the imaging principle of the Lidar sensor, the point cloud closer to the sensor has a larger density, and the remote part is sparse~\cite{codedvtr}. Due to the neighboring aggregation characteristic of the convolution, the predictor tends to output higher results for denser regions and could miss the remote objects (as shown in Fig.~\ref{fig:ablation-density}). To compensate for this bias, we propose density-guided spatial filtering that takes the unique properties of the Lidar point cloud into consideration. Specifically, we use the point cloud BEV density to adjust the predictor score. Therefore, the importance criterion $\mathbf{S}$ is calculated as follows:
\begin{equation}
  \begin{split}
  \mathbf{S} = \mathbf{F}_{score}(\mathbf{X}_{BEV}) = \mathbf{F}_{pred}(\mathbf{X}; \Theta_{pred}) \cdot(\mathbf{D}_{g})^{\beta},
  \end{split}
  \label{equ:density}
\end{equation}
where $D$ is the density heatmap pooled with kernel size of $g$, and $\beta$ is a hyperparameter that tunes the density distribution. The value of $\beta$ is selected for each dataset with the goal of aligning the variance of the predictor score and density distribution on 10 sampled scenes. 
Example in Fig.~\ref{fig:ablation-density} demostrates that the density guidance enlarges the importance score for sparser regions and avoids mistakenly dropping the remote objects.

\subsection{Sparsity Preserving Batch Normalization}
\label{sec:method-SP-BN}

\begin{figure}[h]
    \centering
    \includegraphics[width=0.5\textwidth]{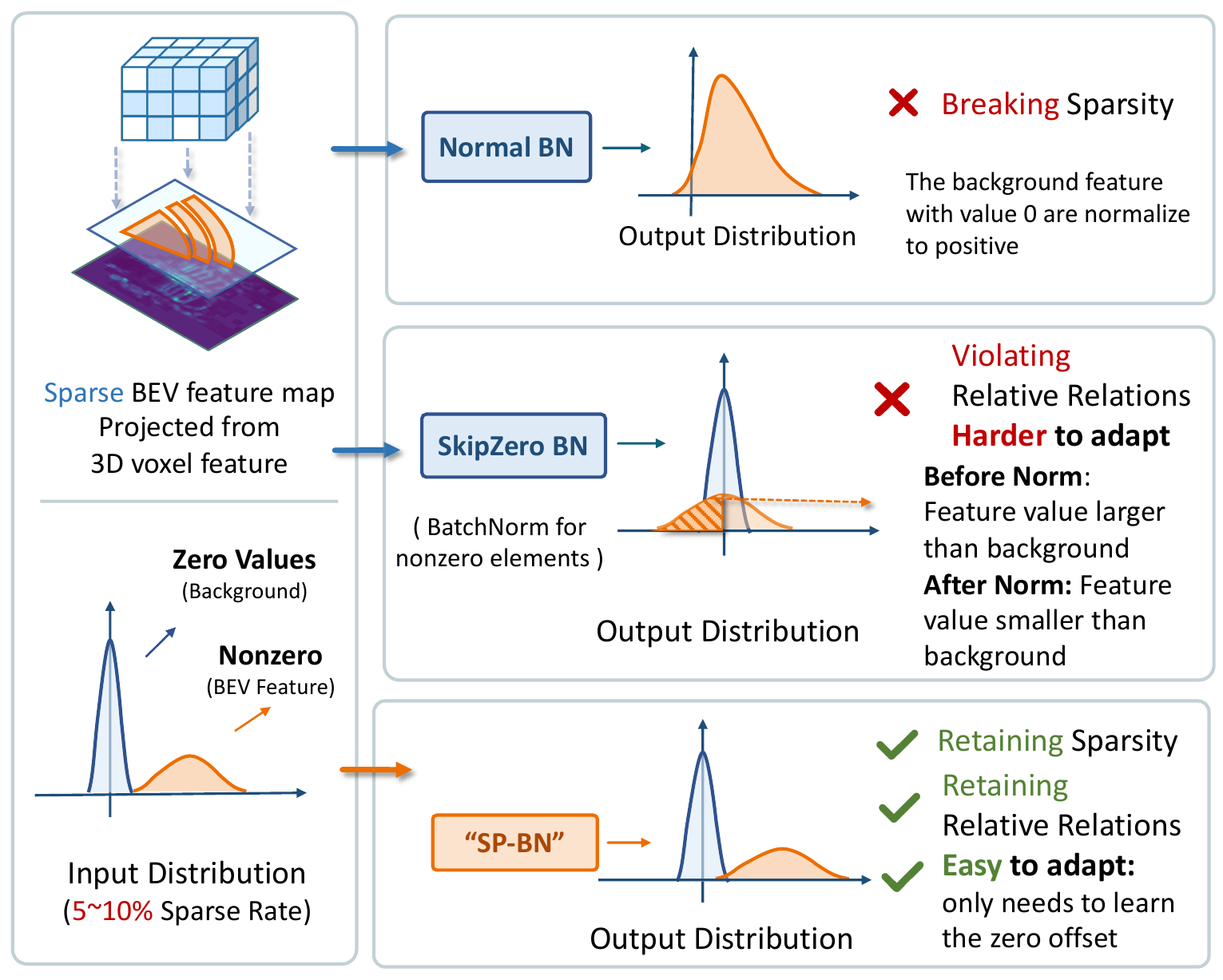}
    \caption{\textbf{Comparison of our proposed sparsity preserving BN with ``Normal BN'' and ``Nonzero BN'' .}}
    \label{fig:sp_bn}
\end{figure}

As illustrated in Fig.~\ref{fig:sp_bn}, the 2D feature map projected from 3D voxel features in the BEV plane is sparse, only 5\% and 20\% features are nonzero for KITTI and nuScenes (the orange part). The rest of the background pixels (the blue ones) are initialized as zero. However, current methods do not utilize such sparsity, and the feature map loses sparsity after the first bacth normalization layer (See Fig.~\ref{fig:oracle}). A large amount of computation and memory is wasted for the ``background'' features with limited information.

A straightforward way to preserve sparsity in 2D BEV backbone is to apply batch normalization only for the nonzero elements.  This approach is described as the ``Nonzero BN'' in Fig.~\ref{fig:sp_bn}. However, we empirically discover that replacing the ``Normal BN'' with ``Nonzero BN'' causes instability in training and moderate performance degradation when fine-tuning from dense pretrained models. We attribute this problem to the violation of the feature's relative relations. As shown in Fig.~\ref{fig:sp_bn}, the orange part with diagonal hatching has larger values than the background features (zero), but after the ``Nonzero BN'', their values are smaller than the background. The finetuning process needs to learn such distribution change thus causing instability. 
To address this problem, we propose a simple but effective modification and introduce the ``Sparsity-preserving BN''. In order to preserve the features' relative relations, the SP-BN leaves out the procedure of subtracting the feature's mean. Therefore, most parts of the nonzero elements remain positive and are distinguishable from the ``background''. The fine-tuning process only needs to learn the offset of the background ``zero'' elements. SP-BN (affine transform omitted) can be formulated as:
\begin{equation}
  \begin{split}
  \hat{x}_i^{(k)} = \frac{x_i^{(k)}}{\sqrt{(\sigma_B^{(k)})^2+\epsilon}},
  \end{split}
  \label{equ:sp_bn}
\end{equation}
where $\sigma_B^{(k)}$ is the standard deviation. Experimental results show that when replacing the normal batch normalization with SP-BN, we could increase the sparsity of 2D BEV heatmap from $0\%$ to $50\%$ without loss of performance. 

\begin{table*}[h]
\centering
\caption{\textbf{Performance comparison of Ada3D and other methods on KITTI $test$ set.} The ``Ada3D-B'' and ``Ada3D-C'' are centerpoint models optimized by Ada3D with different drop rates. }
\label{tab:kitti-test}
\resizebox{1.0\linewidth}{!}{
\begin{tabular}{cccccccccccccc}
\toprule[1pt]
\multirow{2}{*}{\textbf{Mehod}} & \textit{FLOPs} & \textit{Mem} & mAP  & \multicolumn{3}{c}{3D Car (IoU=0.7)} & \multicolumn{3}{c}{3D Ped. (IoU=0.5)} & \multicolumn{3}{c}{3D Cyc. (IoU=0.5)} \\
\cmidrule(lr){5-7} \cmidrule(lr){8-10} \cmidrule(lr){11-13}
 & \textit{Opt.} & \textit{Opt.}  & \textit{(Mod.)} & \textit{Easy} & \textit{Mod.} & \textit{Hard} & \textit{Easy} & \textit{Mod.} & \textit{Hard} & \textit{Easy} & \textit{Mod.} & \textit{Hard} \\
\midrule\midrule
VoxelNet~\cite{voxel_net}     & - & - & 49.05 & 77.47 & 65.11 & 57.73 & 39.48 & 33.69 & 31.50 & 61.22 & 48.36 & 44.37 \\
SECOND~\cite{second}      & - & - & 57.43 & 84.65 & 75.96 & 68.71 & 45.31 & 35.52 & 33.14 & 75.83 & 60.82 & 53.67 \\\
PointPillars~\cite{pointpillar} & - & - & 58.29 & 82.58 & 74.31 & 68.99 & 51.45 & 41.92 & 38.89 & 77.10 & 58.65 & 51.92 \\
SA-SSD~\cite{sa-ssd}       & - & - & - & 88.75 & 79.79 & 74.16 & -  & - & - & - & - & - \\
TANet~\cite{tanet}        & - & - & 59.90 & 84.39 & 75.94 & 68.82 & 53.72 & 44.34 & 40.49 & 75.70 & 59.44 & 52.53 \\
Part-$A^2$~\cite{part-a2}   & - & - & 61.78 & 87.81 & 78.49 & 73.51 & 53.10 & 43.35 & 40.06 & 79.17 & 63.52 & 56.93 \\
SPVCNN~\cite{spvnas}       & - & - & 61.16 & 87.80 & 78.40 & 74.80 & 49.20 & 41.40 & 38.40 & 80.10 & 63.70 & 56.20 \\
\midrule
PointRCNN~\cite{pointrcnn}    & - & - & 57.95 & 86.96 & 75.64 & 70.70 & 47.98 & 39.37 & 36.01 & 74.96 & 58.82 & 52.53 \\  
3DSSD~\cite{3dssd}        & - & - & 55.11 & 87.73 & 78.58 & 72.01 & 35.03 & 27.76 & 26.08 & 66.69 & 59.00 & 55.62 \\
IA-SSD~\cite{ia_ssd}       & - & - & 60.30 & 88.34 & 80.13 & 75.10 & 46.51 & 39.03 & 35.60 & 78.35 & 61.94 & 55.70 \\ 
\midrule
CenterPoint~\cite{centerpoint}  & - & - & 59.96 & 88.21 & 79.80 & 76.51 & 46.83 & 38.97 & 36.78 & 76.32 & 61.11 & 53.62 \\
CenterPoint-Pillar~\cite{centerpoint} & - & - & 57.39 & 84.76 & 77.09 & 72.47 & 44.07 & 37.80 & 35.23 & 75.17 & 57.29 & 50.87  \\
CenterPoint (Ada3D-B)  & \textbf{5.26$\times$} & \textbf{4.93$\times$} & \textbf{59.85} &  87.46 & 79.41 & 75.63 & 46.91 & 39.11 & 36.43 & 76.09 & 61.04 & 53.73 \\
CenterPoint (Ada3D-C)  & \textbf{9.83$\times$} & \textbf{8.49$\times$} & \textbf{57.72} & 82.52 & 74.98 & 69.11 & 43.66 & 38.23 & 34.80 & 75.27 & 59.96 & 52.14 \\
\bottomrule[1pt]
\end{tabular}}
\end{table*}

\begin{table}[h]
\centering
\caption{\textbf{Performance comparison of Ada3D on Nuscenes \textit{val} set.} The ``SPSS-Conv'' model applies pruning for the 3D sparse convolution only, and the ``CenterPoint-0.5W'' uses the 2D backbone with half the usual width.}
\label{tab:nuscens}
\resizebox{0.95\linewidth}{!}{
\begin{tabular}{ccccc}
\toprule
\multirow{2}{*}{Method} & \textit{FLOPs} & \textit{Mem.} & \multirow{2}{*}{mAP} & \multirow{2}{*}{NDS} \\
 & \textit{Opt.} & \textit{Opt.} & & \\
\midrule \midrule
PointPillar~\cite{pointpillar} & - & - & 44.63 & 58.23  \\
SECOND~\cite{second}      & - & - & 50.59 & 62.29 \\
CenterPoint-Pillar~\cite{centerpoint} & - & - & 50.03 & 60.70 \\
\midrule
CenterPoint~\cite{centerpoint}    & \multirow{2}{*}{-} & \multirow{2}{*}{-} & \multirow{2}{*}{55.43} & \multirow{2}{*}{64.63} \\
\small{\textit{(voxel=0.1)}} & & & & \\
CenterPoint-Ada3D & \multirow{2}{*}{2.32$\times$} & \multirow{2}{*}{2.61$\times$} &  \multirow{2}{*}{54.80} & \multirow{2}{*}{63.53} \\
\small{\textit{(voxel=0.1)}} & & & & \\
\midrule
CenterPoint~\cite{centerpoint}  & \multirow{2}{*}{-} & \multirow{2}{*}{-} & \multirow{2}{*}{59.22} & \multirow{2}{*}{66.48} \\
\small{\textit{(voxel=0.075)}} & & & & \\
SPSS-Conv~\cite{sps-conv}  & \multirow{2}{*}{1.14$\times$} & \multirow{2}{*}{1.14$\times$} & \multirow{2}{*}{57.80} & \multirow{2}{*}{65.69} \\
\small{\textit{(voxel=0.075)}} & & & & \\
CenterPoint-0.5W~\cite{centerpoint}  & \multirow{2}{*}{2.78$\times$} & \multirow{2}{*}{2.78$\times$} & \multirow{2}{*}{57.19} & \multirow{2}{*}{64.08} \\
\small{\textit{(voxel=0.075)}} & & & & \\
CenterPoint-Ada3D  & \multirow{2}{*}{3.34$\times$} & \multirow{2}{*}{3.96$\times$} & \multirow{2}{*}{58.62} & \multirow{2}{*}{65.68} \\
\small{\textit{(voxel=0.075)}} & & & & \\
\midrule
VoxelNeXT~\cite{voxelnext} & - & - & 60.50 & 66.60 \\
VoxelNeXT-Ada3D~\cite{voxelnext} & 1.19$\times$ & 1.20$\times$ & 59.75 & 65.84 \\
\midrule
\bottomrule
\end{tabular}}
\end{table}

\begin{table}[h]
\centering
\caption{\textbf{Performance comparison of Ada3D on ONCE \textit{val} set.} The results are taken from the OpenPCDet~\cite{openpcdet} implementation.}
\label{tab:once}
\resizebox{1.0\linewidth}{!}{
\begin{tabular}{ccccccc}
\toprule
\multirow{2}{*}{Method} & \textit{FLOPs} & \textit{Mem.} & \multirow{2}{*}{mAP} & \multirow{2}{*}{Veh.} & \multirow{2}{*}{Ped.} & \multirow{2}{*}{Cyc} \\
 & \textit{Opt.} & \textit{Opt.} & & & & \\
\midrule 
PointRCNN~\cite{pointrcnn} & - & - & 28.74 & 52.09 & 4.28 & 29.84 \\
PointPillar~\cite{pointpillar} & - & - & 44.34 & 68.57 & 17.63 & 46.81  \\
SECOND~\cite{second}      & - & - & 51.89 & 71.16 & 26.44 & 58.04 \\
PVRCNN~\cite{pvrcnn}  & - & - & 53.55 & 77.77 & 23.50 & 59.37 \\
CenterPoint~\cite{centerpoint}  & - & - & 63.99 & 75.69 & 49.80 & 66.48 \\
\midrule
CenterPoint & \multirow{2}{*}{2.32$\times$} & \multirow{2}{*}{2.61$\times$} &  \multirow{2}{*}{62.68} & \multirow{2}{*}{73.43} & \multirow{2}{*}{49.09} & \multirow{2}{*}{65.53} \\
\small{\textit{(Ada3D)}} & & & & \\
\bottomrule
\end{tabular}}
\end{table}

\section{Experiments}
\label{sec:exp}


\subsection{Implemention Details}
\label{sec:implemention-detail}

\textbf{KITTI and nuScenes and ONCE dataset}
The KITTI dataset has 7481 training images and 7518 test images with corresponding point clouds. The object to detect have 3 classes: car, pedestrian, and cyclist, the boxes are classified into three subsets: “Easy”, “Moderate” and “Hard” based on the levels of difficulty. The detection results are evaluated by average precision (AP) for each subset with IoU threshold 0.7 for cars and 0.5 for pedestrians and cyclists. 
The nuScenes dataset comprises 1000 driving sequences with annotations in the form of bounding boxes for 10 object classes. 
The commonly used metrics are the mean Average Precision (mAP) and the nuScenes detection score (NDS).
NDS is the weighted average of mAP and other box characteristics, such as translation and orientation.
The ONCE~\cite{once} dataset provides Lidar point clouds collected from downtown and suburban areas of multiple cities for 3D object detection.
For supervised training, the training set contains 5k labelled scenes and the validation set contains 3K scenes. The commonly used mAP is adopted as the evaluation metric. 

\textbf{Adaptive inference design}
 We apply Ada3D to CenterPoint~\cite{centerpoint} model on both datasets. Due to the original CenterPoint paper does not conduct experiments on KITTI, we follow the author's released code~\cite{centerpoint_kitti} to construct the CenterPoint model on KITTI. We replace all the batch normalization layers in the 2D backbone with sparsity-preserving BN. We apply adaptive inference at the 2nd and 4th layer of the 3D and 2D backbone. The $R_{drop}$ is a hyperparameter (e.g., 25\%/50\%) to control how many features to drop. The predictor's input resolution is set as the scene size divided by voxel size$\times$8 . Max poolings and same padding upsample layers are adopted to align features of different sizes. 
The predictor consists of 3 convolution layers with channel size $\left[16,32,16\right]$ and group size of 8. The predictor is trained with adam optimizer with one-cycle learning rate shceduling~\cite{one_cycle} of learning rate 0.003 for 10 epochs. To recover the performance, we adopt an interleaved scheme that alternates between finetuning the model with adaptive inference for 5 (2 for nuScenes) epochs and training the predictor for 1 epoch and repeat this process for a total of 5 times. The $\sigma$ for ground-truth heatmap is 5.0. The density guidance $\beta$ is set as 0.5 and 0.7 for KITTI and nuScenes/ONCE. 

\textbf{Hardware experiments settings} We measure the latency and memory usage of convolution layers on an Nvidia RTX 3090 GPU using CUDA 11.1. We implemented sparse convolution operations using the gather-GEMM-scatter dataflow in TorchSparse v2.0.0~\cite{torchsparse} and SpConv v.2.2.6~\cite{sparseconvnet}. 
To measure latency, we synchronized the GPU and recorded the starting and ending times. To measure peak memory usage, we embedded the PyTorch Memory Utils~\cite{pytorch_memory_profile} into the engine frontend.


\begin{table*}[h]
\centering
\caption{\textbf{Ablation studies and quantitve efficiency improvements of different Ada3D models on KITTI \textit{val}.} ``IP'' stands for ``importance predictor'', ``DG'' for ``density-guided spatial filtering'', ``SP-BN'' for ``sparsity preserving batch normalization''. The ``FLOPs'' and ``Mem.'' calculates the normalized resource consumption of the optimized model.}
\label{tab:ablation-eff}
\resizebox{1.0\linewidth}{!}{
\begin{tabular}{cccccccccccc}
\toprule
\multirow{2}{*}{Method} & \multicolumn{3}{c} {\textit{Technique}} & \multicolumn{2}{c}{\textit{FLOPs}} & \multicolumn{2}{c}{\textit{Mem.}} & mAP & Car Mod. & Ped. Mod. & Cyc. Mod. \\
\cmidrule(lr){2-4} \cmidrule(lr){5-6}  \cmidrule(lr){7-8}
 & IP & DG & SP-BN &  \textit{3D} & \textit{2D} & \textit{3D} & \textit{2D} & (Mod.) & (IoU=0.7) & (IoU=0.5) & (IoU=0.5)  \\
\midrule \midrule
CenterPoint & - & - & - & 1.00 & 1.00 & 1.00 & 1.00 & 66.1 & 79.4 (-) & 53.4 (-) & 65.5 (-) \\
CenterPoint (SP-BN) & - & - & \checkmark & 1.00 & 0.49 & 1.00 & 0.45 & 66.0 & 79.1 (-0.3) & 53.3 (-0.1) & 65.6 (+0.1)  \\
CenterPoint (Ada3D-A) & \checkmark & \checkmark & \checkmark & 1.00 & 0.22 & 1.00 & 0.25 & 66.4 & 79.5 (+0.1) & 53.6 (+0.2) & 66.1 (+0.6) \\
CenterPoint (Ada3D-B) & \checkmark & \checkmark & \checkmark & 0.66 & 0.18 & 0.68 & 0.17 & 66.1 & 79.1 (-0.3) & 54.0 (+0.6) & 65.3 (-0.3) \\
CenterPoint (Ada3D-B w.o. DG) & \checkmark & - & \checkmark & 0.64 & 0.18 & 0.66 & 0.16 & 65.1 & 78.8 (-0.6) & 51.6 (-1.8) & 64.9 (-0.6)  \\
CenterPoint (Ada3D-C) & \checkmark & \checkmark & \checkmark & 0.39 & 0.08 & 0.43 & 0.07 & 65.4 & 77.6 (-1.8) & 53.5 (+0.2) & 65.1 (-0.4) \\
\bottomrule
\end{tabular}
}
\end{table*}

\begin{figure}[ht]
    \centering
    \includegraphics[width=0.5\textwidth]{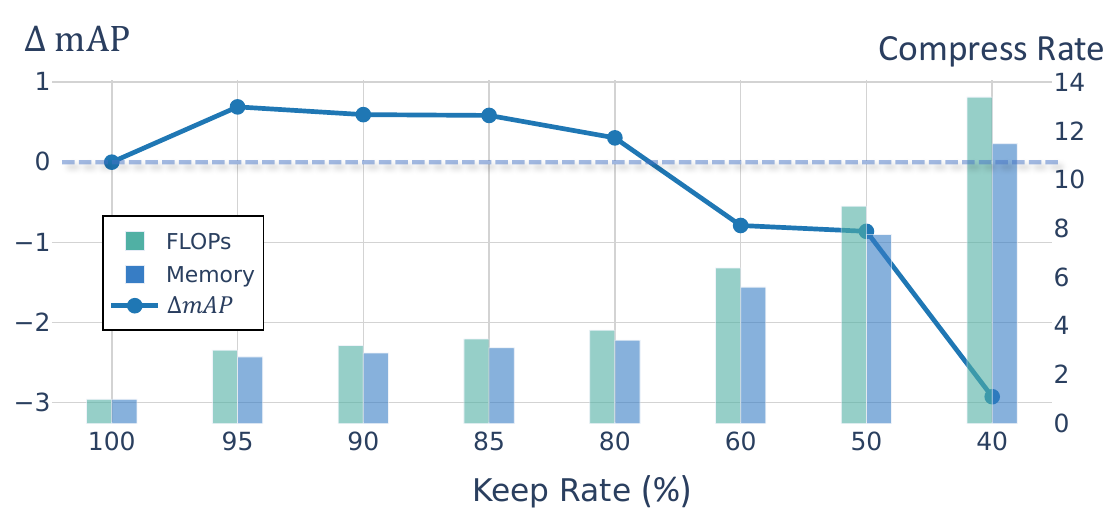}
    \caption{\textbf{Ada3D's Performance under different $R_\text{drop}$.} Left: the relative mAP compared with baseline without Ada3D. Right: The FLOPs and memory compress rate.}
    \label{fig:perf_bars}
\end{figure}

\subsection{Performance and Efficiency Comparison}
\label{sec:perf-eff}

We first present the performance and resource consumption of Ada3D optimized model on KITTI and nuScenes. We estimate the memory cost of the model by summing the intermediate activation sizes following recent literature~\cite{hardware_survey}. As could be seen from Table.~\ref{tab:kitti-test}, \textbf{the Ada3D optimized model achieves comparable performance with other methods of different paradigms while compressing the model's computational and memory cost.} In Table.~\ref{tab:ablation-eff} and Fig.~\ref{fig:perf_bars}, we present Ada3D model with different drop rates. The model size could be effectively tuned with the drop rate $R_{\text{drop}}$ to fit different resource budgets. ``Ada3D-A'' model only conducts adaptive inference for 2D backbone, it improves the model performance while reducing the dense rate of BEV features from 100\% to 20\% . ``Ada3D-B'' model reduces 40\% 3D voxels and more than 80\% 2D pixels and compresses the computaional and memory cost of the model by 5$\times$ without performance degradation. ``Ada3D-C'' model reduces 60\% 3D voxels and more than 90\% 2D pixels with moderate performance loss, and reduces the model's computation and memory cost by an order of magnitude. Table.~\ref{tab:nuscens} presents the performance on nuScenes, Ada3D optimized CenterPoint model achieves 2$\sim$4$\times$ FLOPs and memory savings with less than 1\% performance drop. 
\textbf{Compared with methods that focus on reducing the model redundancy} (``CenterPoint-0.5W'' and ``SPSS-Conv''), \textbf{Ada3D achieves a larger compression rate with less performance drop}.

\subsection{Hardware Experiments}
\label{sec:hardware}

\begin{figure}[h]
    \centering
    \includegraphics[width=0.5\textwidth]{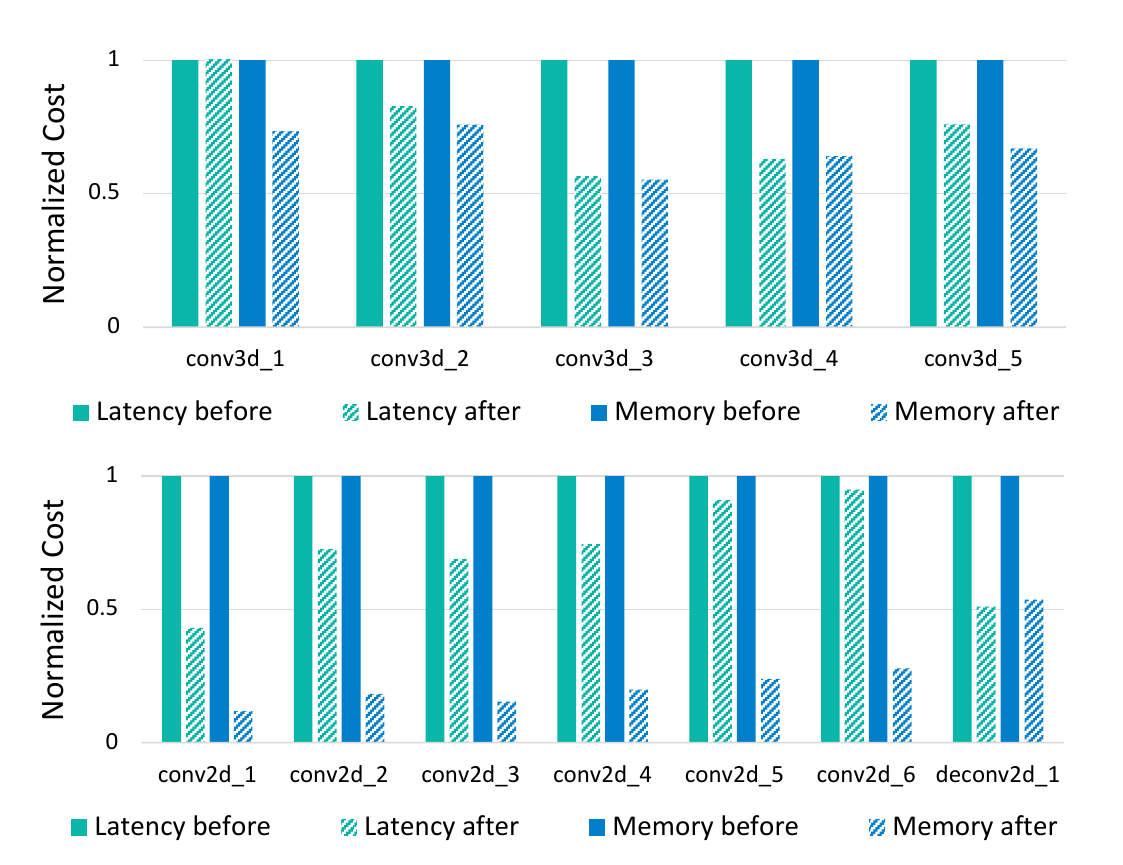}
    \caption{\textbf{Layer-wise GPU latency and peak memory optimization for Ada3D-B}. The green and blue bars stand for the latency and peak memory. The diagonal filled bars are Ada3D optimized costs.}
    \label{fig:hardware_op}
\end{figure}

\begin{figure}[h]
    \centering
    \includegraphics[width=0.5\textwidth]{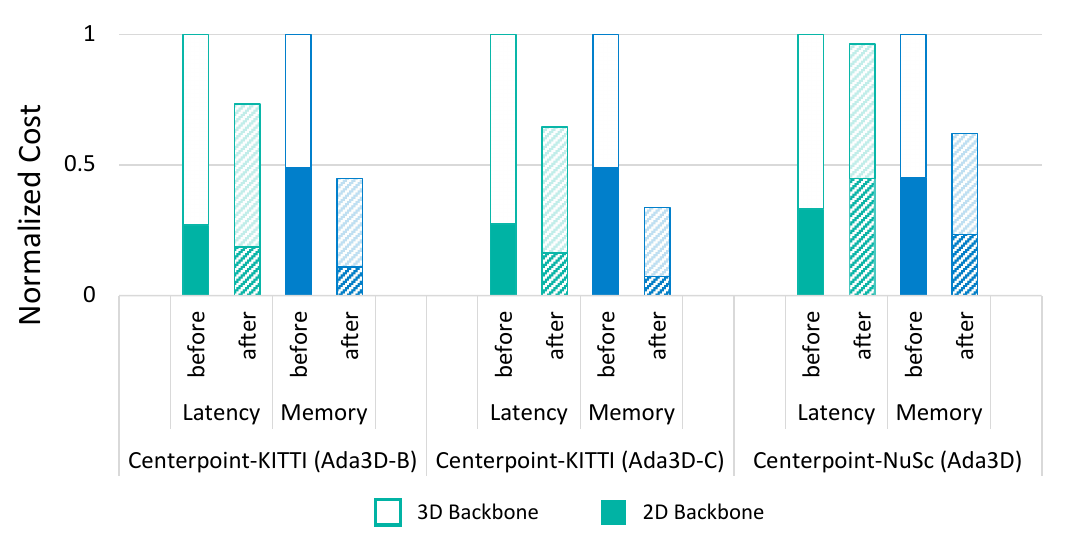}
    \caption{\textbf{End-to-end GPU latency and peak memory optimization for Ada3D}. The green and blue bars stand for the latency and peak memory cost respectively. The filled/unfilled bars represent the 2D/3D backbone. }
    \label{fig:hardware_end2end}
\end{figure}

We conduct hardware profiling of the Ada3D model using sparse convolution GPU libraries~\cite{torchsparse,sparseconvnet}. Fig.\ref{fig:hardware_op} illustrates the reduction of GPU latency and peak memory for each layer, while Fig.\ref{fig:hardware_end2end} presents the end-to-end hardware specs for the 3D and 2D backbones, respectively. From the results, we draw the following conclusions. First, \textbf{by using SP-BN and spatial filtering, we retain high sparsity of the 2D feature map}, which brings significant reductions in peak memory and computation for the 2D backbone. For instance, the "conv2d\_1" layer shows a 2.5$\times$ latency and 8.5$\times$ memory improvement, and the overall memory of the 2D backbone is reduced by 4.5$\times$, 6.7$\times$, and 1.9$\times$ for each model. Second, \textbf{the end-to-end latency of the 3D backbone aligns with the drop rate.} The latency for the 3D backbone is 0.74$\times$, 0.56$\times$, and 0.77$\times$ of the pre-optimized ones, which corresponds to the drop rate (25\%, 50\%, 25\%). Third, \textbf{Ada3D is more effective for larger scenes and finer voxel sizes}, since there exists more potential for exploiting the spatial sparsity.
For the nuScenes Ada3D model, only peak memory optimization is achieved, but the latency remains similar. This is because that due to resource constraints, a larger voxel size is often used at the cost of inferior performance~\cite{spvnas}. The nuScenes BEV feature map is processed in a relatively low resolution ($\left[128,128\right]$), thus the dense rates of the deeper layer's feature maps remain high, and using sparse convolution to process them takes longer than normal convolution. Future directions to improve this include further reducing redundancy or adopting more hardware acceleration techniques. Addtionaly, improving the efficiency could enable finer voxel size, which could in turn enhances performance and safety for safety-critic autonomous driving applicaiton.

\section{Analysis and Discussions}
\label{sec:analysis}

\subsection{Ablation Studies}
\label{sec:ablations}

\begin{figure}[h]
    \centering
    \includegraphics[width=0.95\linewidth]{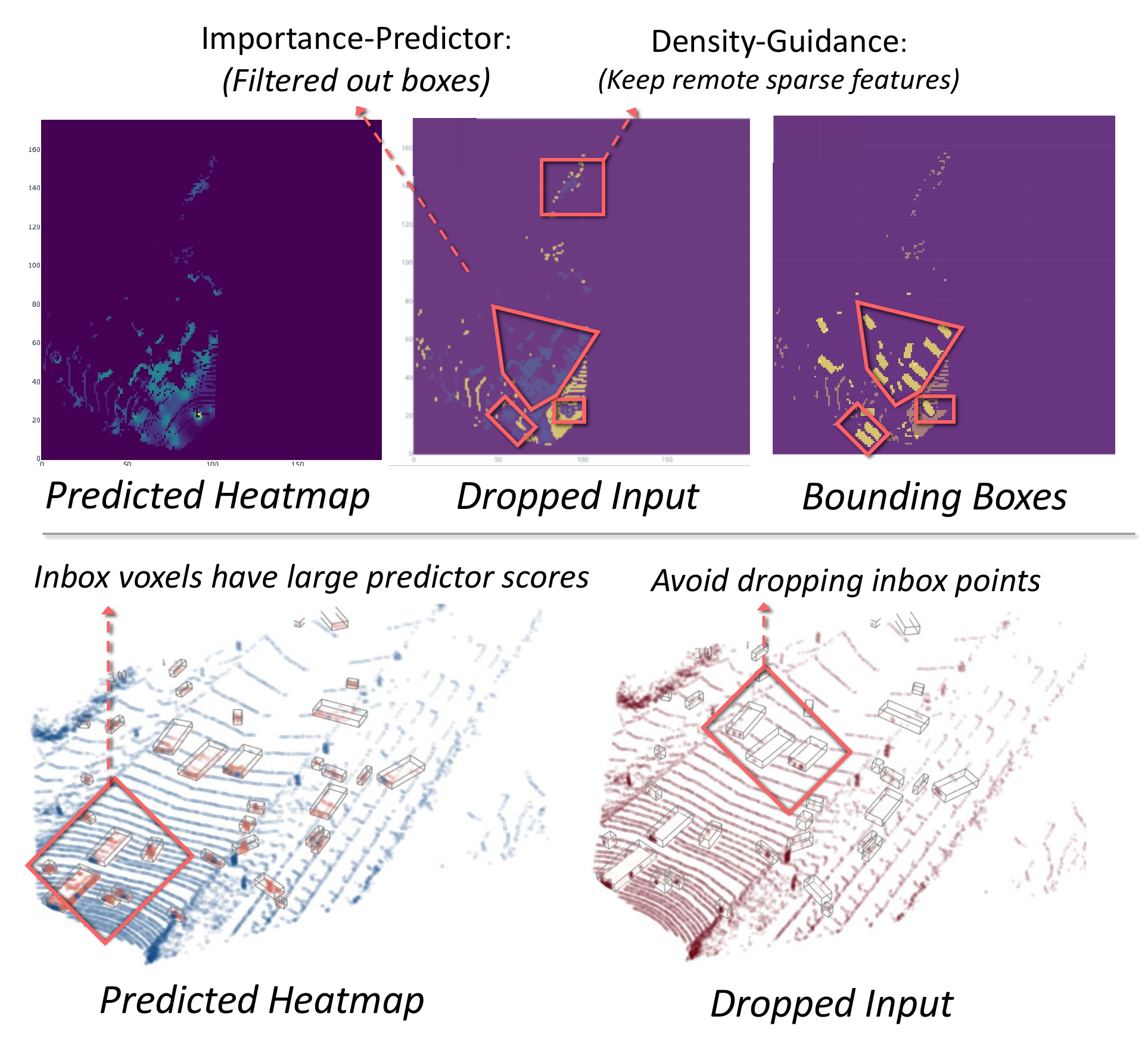}
    \caption{\textbf{Qualitative Results for 3D and 2D adaptive inference on KITTI and nuScenes dataset.} Visualization of the predicted heatmap and dropped input in BEV and voxel space. The predictor identifies the voxels/pixels inside the bounding box and avoids dropping them.}
    \label{fig:qualitative}
\end{figure}

\begin{figure}[h]
    \centering
    \includegraphics[width=0.95\linewidth]{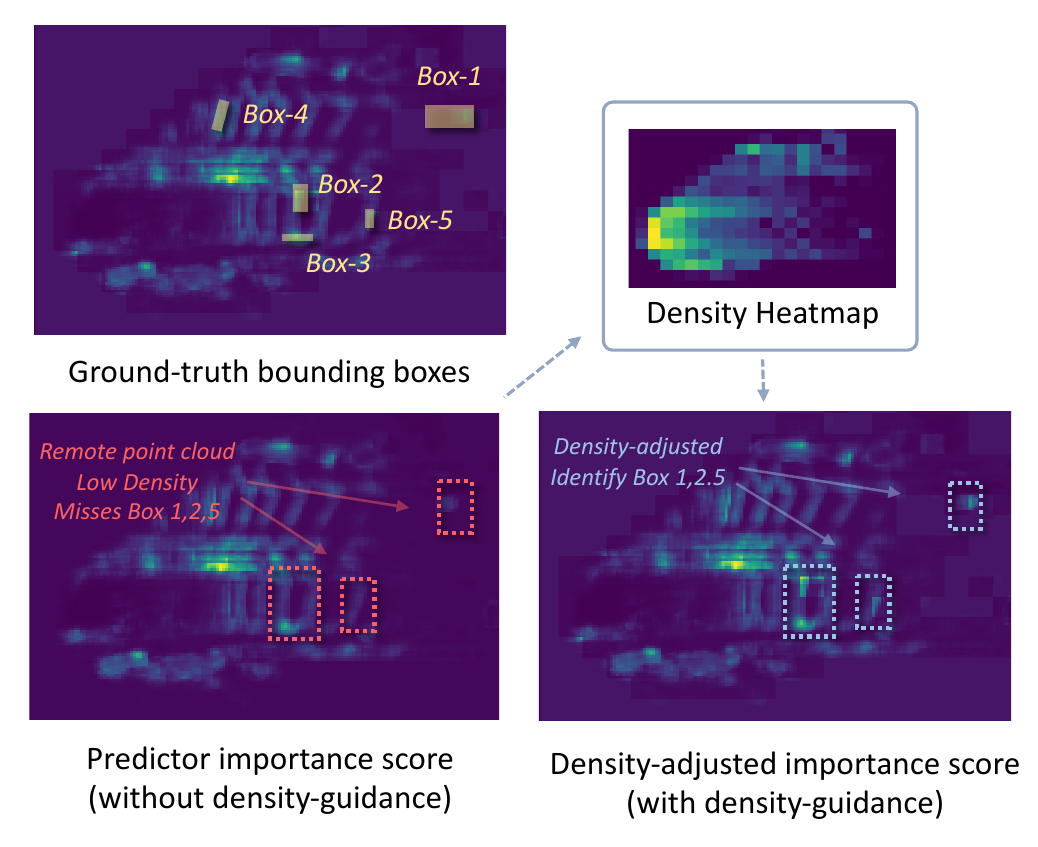}
    \caption{\textbf{Example of the density guidance corrects the drop of smaller remote objects.} Visualization of the predictor scores and density-guided scores.} 
    \label{fig:ablation-density}
\end{figure}

\textbf{Importance predictor accurately evaluates the input features' importance.} In Table.~\ref{tab:ablation-eff}, comparing ``Ada3D-A'' and ``SP-BN'', the predictor increases the 2D feature map's sparsity from 50\% to 80\% upon SP-BN. As shown in Table.~\ref{tab:ablation-criterion}, Among the least important 25\%/50\% predicted, only 1.5\%/7.8\% features are mistakenly evaluated. Fig.~\ref{fig:ablation-density} and Fig.~\ref{fig:qualitative} present the visualization of predictor heatmaps in both the BEV and voxel space. The predictor recognizes features within the box and avoids dropping them. 

\textbf{Density guidance avoids dropping the remote small-sized objects.} In Table.~\ref{tab:ablation-eff}, comparing the ``Ada3D-B'' models with and without density guidance, simply using predictor scores causes notable performance degradation, especially for the pedestrian (-2.4\%) with smaller sizes. Fig.~\ref{fig:ablation-density} shows the example of density guidance correcting the drop of remote small objects. The predictor fails to correctly detect features for box-1,2,5 due to low density, and the density guidance compensates for such error. We also compare different importance criteria under different drop rates $R_{drop}$ in Table.~\ref{tab:ablation-criterion}. The $R_{inbox}$ denotes the percentage of dropped features that are in the ground-truth bounding box. Solely using the predictor score (IP) or density (DG) results in high $R_{inbox}$ and performance degradation.

\textbf{SP-BN preserves the sparsity without performance drop.} Table.~\ref{tab:ablation-eff} shows that introducing the SP-BN increases the sparsity of 2D BEV features from 0\% to 50\% with no performance drop. Using the ``Noraml BN'' sacrifices the sparsity. Additionally, adopting the ``Nonzero BN'' for the entire network results in notable performance loss when finetuning from pretrained dense backbone. We hypothesize that it is because of the ``Nonzero-BN'' needs to learn the entire distribution shift, while the ``SP-BN'' only needs to learn the offset of zero elements. 

\begin{table}[h]
\centering
\caption{\textbf{Performance of combining Ada3D and model-level compression method.}}
\label{tab:spvnas}
\resizebox{1.01\linewidth}{!}{
\begin{tabular}{ccccccc}
\toprule
\multirow{2}{*}{Method} & \textit{FLOPs} & \textit{Mem.} & \multirow{2}{*}{mAP} & \multicolumn{3}{c}{KITTI Mod.} \\
\cmidrule(lr){5-7}
 & \textit{Opt.} & \textit{Opt.} & & Car. & Ped.  & Cyc. \\
\midrule 
CenterPoint~\cite{centerpoint}  & - & - & 66.1 & 79.4 & 53.4 & 65.5 \\
\midrule
CenterPoint & \multirow{2}{*}{1.07$\times$} & \multirow{2}{*}{1.07$\times$} &  \multirow{2}{*}{65.5} & \multirow{2}{*}{79.2} & \multirow{2}{*}{52.1} & \multirow{2}{*}{65.3} \\
\small{\textit{(SPVNAS)}} & & & & \\
\midrule
CenterPoint & \multirow{2}{*}{3.95$\times$} & \multirow{2}{*}{4.35$\times$} &  \multirow{2}{*}{65.5} & \multirow{2}{*}{78.6} & \multirow{2}{*}{52.5} & \multirow{2}{*}{65.5} \\
\small{\textit{(SPVNAS+Ada3D)}} & & & & \\
\bottomrule
\end{tabular}}
\end{table}

\begin{table}[h]
\centering
\caption{\textbf{Comparison of adopting different importance criteria for input spatial filtering.} ``IP'' and ``DG'' stand for importance predictor and density guidance.The ``$R_{\text{drop}}$'' represents the drop ratio. ``$R_{\text{inbox}}$'' represents the percentage of dropped inputs that are within the ground truth bounding box (the lower the better).}
\label{tab:ablation-criterion}
\resizebox{1.0\linewidth}{!}{
\begin{tabular}{cccccccc}
\toprule
\multicolumn{2}{c}{\textbf{$f_{\text{score}}$}} & \multirow{2}{*}{\textbf{$R_{\text{drop}}$}} & \multicolumn{2}{c}{\textbf{$R_{\text{inbox}}$}} & \multicolumn{3}{c}{KITTI Mod. AP} \\
\cmidrule(lr){1-2} \cmidrule(lr){4-5} \cmidrule(lr){6-8}
\textbf{IP} & \textbf{DG} &  & \textit{3D} & \textit{2D} & Car. & Ped. & Cyc. \\
\midrule\midrule
- & - & - & - & - & 79.1 & 53.3 & 65.6 \\
- & \checkmark &  25\% & 12.3\% & 9.4\% & 76.4 & 45.6 & 59.4\\
\checkmark & - & 25\% & 1.4\% & 1.1\% & 78.8 & 51.6 & 64.9 \\
\checkmark & \checkmark & 25\% & \textbf{0.8\%} & \textbf{0.0\%} & \textbf{79.1} & \textbf{54.0} & \textbf{65.2} \\
\midrule
- & \checkmark &  50\% & 17.6\% & 20.3\% & 72.1 & 39.4 & 55.6 \\
\checkmark & - & 50\% & 6.8\% & 8.8\% & 76.9 & 50.2 & 63.7 \\
\checkmark & \checkmark & 50\% & \textbf{5.2\%} & \textbf{7.5\%} & \textbf{77.6} & \textbf{53.5} & \textbf{65.1} \\
\bottomrule
\end{tabular}}
\end{table}

\begin{table}[h]
\centering
\caption{\textbf{Comparison of different BN types.} SP-BN maintains both performance and sparsity.}
\label{tab:ablation-bn}
\resizebox{0.8\linewidth}{!}{
\begin{tabular}{ccccc}
\toprule
\multirow{2}{*}{BN Type} & \multirow{2}{*}{Sparse} & \multicolumn{3}{c}{KITTI Mod. AP} \\
\cmidrule(lr){3-5}
 &  & Car. & Ped. & Cyc. \\
\midrule
Normal BN & - & 79.4 & 53.4 & 65.5 \\
Without BN & \checkmark & 76.3 & 43.5 & 49.7 \\
Nonzero BN & \checkmark & 74.5 & 39.4 & 47.3 \\
SP-BN & \checkmark & \textbf{79.1} & \textbf{53.3} & \textbf{65.6}  \\

\bottomrule
\end{tabular}}
\end{table}


\subsection{Analysis of the Adaptive Inference}
\label{sec:adaptive-inference-analysis}


\textbf{Ada3D introduces negligible overhead.} The extra cost that Ada3D introduces is the predictor inference. The predictor is conducted in a relatively low resolution and utilizes group convolution. The predictor's computational cost is less than 1\% of the 2D BEV backbone, which is negligible. The training cost of Ada3D includes a brief training of the predictor and model finetuning, which accounts for less than 30\% of the original model's training time. 

\textbf{Ada3D could improve the performance} Adaptive inference removes redundant input features and saves computation and memory costs. However, adaptive inference does not necessarily have negative effects on performance. As shown in Table.~\ref{tab:ablation-eff} and Fig.~\ref{fig:perf_bars}, ``Ada3D-A'' improves the performance. We infer that the dropped redundant part is noisy and has negative effects on the training process. 

\textbf{Ada3D could be applied on the fully sparse 3D detectors.} Fully sparse detectors (e.g., FSD~\cite{FSD}, FSD++~\cite{FSD++}, VoxelNeXt~\cite{voxelnext}) eliminate the dense BEV feature with novel architecture designs that directly process the sparse BEV feature to generate boxes. These models can still benefit from Ada3D's spatial filtering, which further reduces redundant inputs. As shown in Tab.~\ref{tab:nuscens}, when applying Ada3D for VoxelNeXT model, we further reduce 20\% of redundant voxels with moderate performance degradation. 

\textbf{Ada3D could work on par with the model-level compression method and further improve efficiency.} In comparison with existing model-level compression methods, Ada3D takes the perspective of compressing the spatial redundancy. Therefore, Ada3D could be  combined with existing model-level to further improve efficiency. We adapt the SPVNAS~\cite{spvnas} searched model to the 3D backbone of Centerpoint, and employ Ada3D to further compress it. As seen in Table.~\ref{tab:spvnas}, Ada3D could further reduce the computaional and memory cost of SPVNAS optimized model.



\begin{figure}[h]
    \centering
    \includegraphics[width=0.95\linewidth]{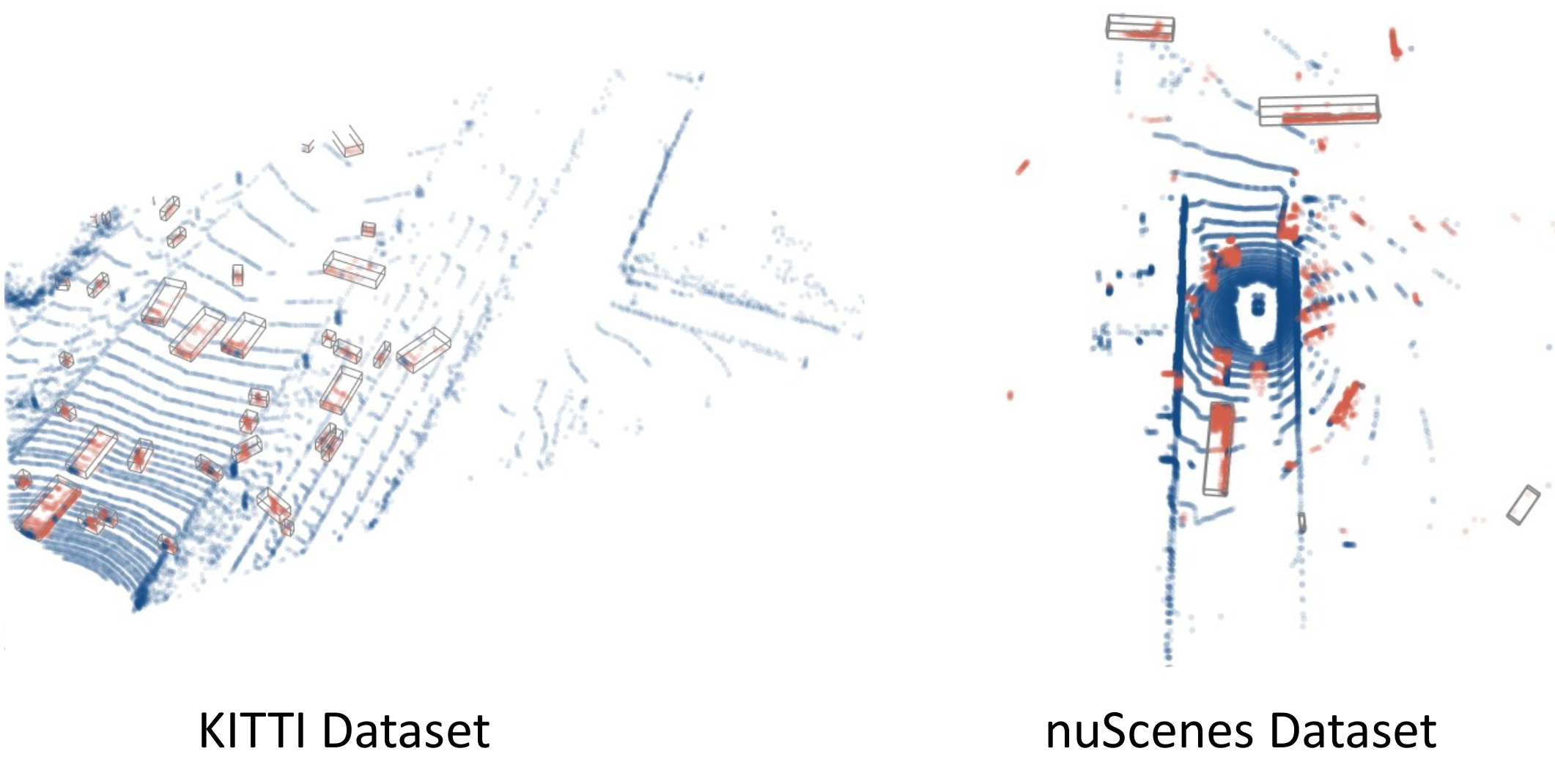}
    \caption{\textbf{Visualization of the 3D point cloud colored by predicted importance score.} The orange represents higher importance, and the blue stands for the lower.} 
    \label{fig:score-vis}
\end{figure}

\section{Limitations and Future Directions}

The 2D BEV backbone exhibits only moderate latency improvement at relatively low sparse rates (e.g., 30\%$\sim$50\%). Further exploration of higher sparsity and hardware designed tailored for utilizing the existing sparsity is necessary. Additionally, the Ada3d optimized model shows moderate performance decay with plain finetuning for recovery. To further enhance its performance, more advanced tuning techniques such as distillation could be employed. Additionally, we could extend the usage of Ada3D to more 3D detectors and other tasks. 



\section{Acknowledgement}
This work was supported by National Natural Science Foundation of China (No. U19B2019, 61832007), Tsinghua University Initiative Scientific Research Program, Beijing National Research Center for Information Science and Technology (BNRist), Tsinghua EE Xilinx AI Research Fund, and Beijing Innovation Center for Future Chips.

{\small
\bibliographystyle{ieee_fullname}
\bibliography{egbib}
}


\clearpage

\appendix


\section{Qualitative results}
\label{sec:qualitative_more}

\begin{figure*}[h]
    \centering
    \includegraphics[width=0.99\textwidth]{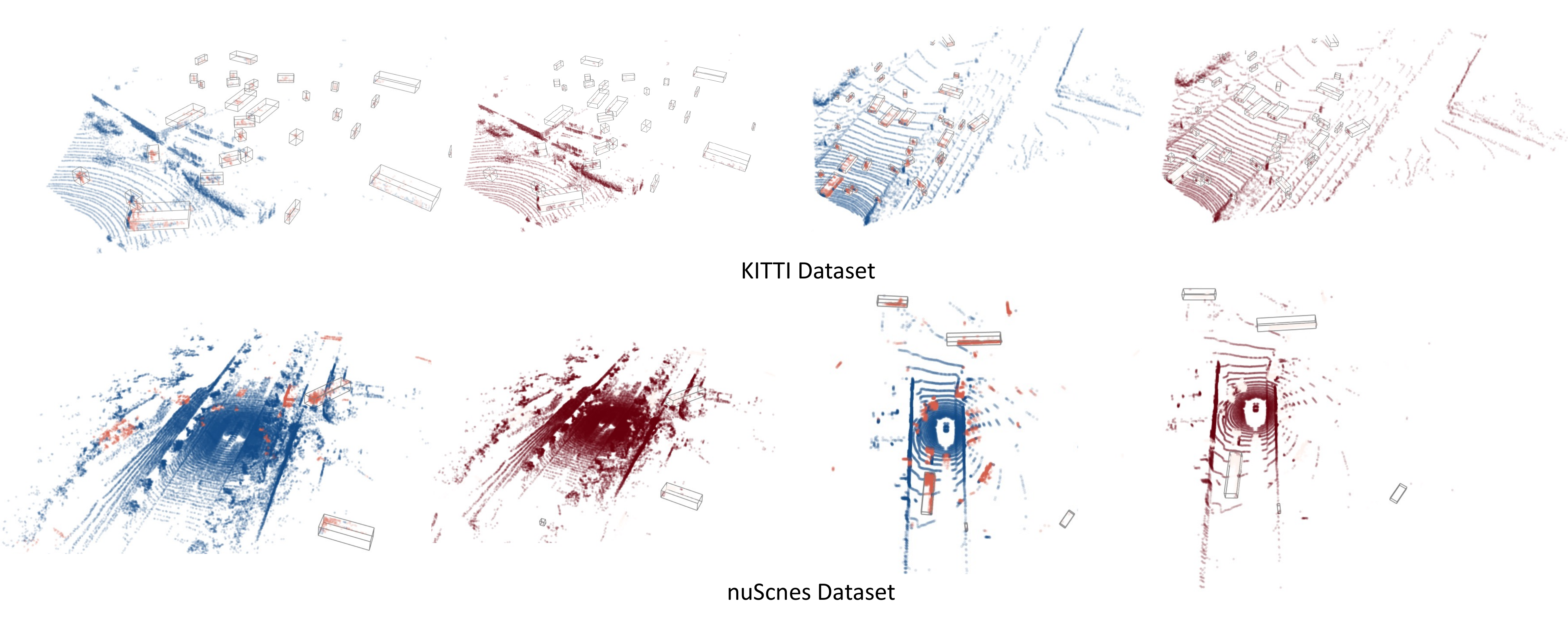}
    \caption{\textbf{Visualization of the 3D point cloud colored by predicted importance score for different datasets.} The orange represents higher importance, and the blue stands for the lower. The red points stand for dropped points given a high drop rate.}
    \label{fig:qualitative_more}
\end{figure*}

We present additional qualitative results on KITTI and nuScenes datasets in Fig.~\ref{fig:qualitative_more}. The Ada3D model gives lower importance prediction for background points (e.g., road plane and building) and over-dense points closer to the Lidar. When using a very large drop rate, Ada3D still avoids dropping the points within the ground-truth bounding boxes. 

\section{Discussions}

\textbf{Effects of sparsity-preserving batch normalization.} Prior studies commonly treat the 2D BEV feature map as dense and apply standard convolution on it, Ada3D introduces the sparsity-preserving batch normalization to maintain the sparsity of the BEV features and adopt sparse convolution for acceleration. It is worth noting that using the ``Nonzero BN'' together with the ``gather-scatter'' sparse convolution could also maintain sparsity for BEV features. 
Nonetheless, we empirically find that it may lead to performance degradation, and our proposed sparsity-preserving BN mitigates the performance decay.   

\textbf{Ada3D could work on par with the model-level compression method and further improve efficiency.} In comparison with existing model-level compression methods, Ada3D takes the perspective of compressing the spatial redundancy. Therefore, Ada3D could be  combined with existing model-level to further improve efficiency. We adapt the SPVNAS~\cite{spvnas} searched model to the 3D backbone of Centerpoint, and employ Ada3D to further compress it. As seen in Table.~\ref{tab:spvnas}, Ada3D could further reduce the computaional and memory cost of SPVNAS optimized model.


\begin{table}[h]
\centering
\caption{\textbf{Performance of combining Ada3D and model-level compression method.}}
\label{tab:spvnas}
\resizebox{1.01\linewidth}{!}{
\begin{tabular}{ccccccc}
\toprule
\multirow{2}{*}{Method} & \textit{FLOPs} & \textit{Mem.} & \multirow{2}{*}{mAP} & \multicolumn{3}{c}{KITTI Mod.} \\
\cmidrule(lr){5-7}
 & \textit{Opt.} & \textit{Opt.} & & Car. & Ped.  & Cyc. \\
\midrule 
CenterPoint  & - & - & 66.1 & 79.4 & 53.4 & 65.5 \\
\midrule
\small{\textit{(SPVNAS)}} & & & & \\
\midrule
CenterPoint & \multirow{2}{*}{3.95$\times$} & \multirow{2}{*}{4.35$\times$} &  \multirow{2}{*}{65.5} & \multirow{2}{*}{78.6} & \multirow{2}{*}{52.5} & \multirow{2}{*}{65.5} \\
\small{\textit{(SPVNAS+Ada3D)}} & & & & \\
\bottomrule
\end{tabular}}
\end{table}

\textbf{Ada3D could enable finer voxel size and improve performance.}
Due to the resource constraints, a larger voxel size is often used to reduce the model cost, but this results in inferior performance~\cite{torchsparse}. As shown in Table.~\ref{tab:voxel_size}, compared to using a larger voxel size (0.1). Ada3D could enable finer voxel size (0.075) and improve model performance with smaller cost. 


\begin{table}[h]
\centering
\caption{\textbf{Performance comparison of Ada3D and enlarging the voxel size on Nuscenes \textit{val} set.} The ``FLOPs Size'' and ``Mem. Size'' stands for the flops and memory cost normalized by the centerpoint model with voxel size 0.075.}
\label{tab:voxel_size}
\resizebox{0.95\linewidth}{!}{
\begin{tabular}{ccccc}
\toprule
\multirow{2}{*}{Method} & \textit{FLOPs} & \textit{Mem.} & \multirow{2}{*}{mAP} & \multirow{2}{*}{NDS} \\
 & \textit{Size} & \textit{Size} & & \\
\midrule \midrule
CenterPoint  & \multirow{2}{*}{1.00$\times$} & \multirow{2}{*}{1.00$\times$} & \multirow{2}{*}{59.22} & \multirow{2}{*}{66.48} \\
\small{\textit{(voxel=0.075)}} & & & & \\
\midrule
CenterPoint    & \multirow{2}{*}{0.70$\times$} & \multirow{2}{*}{0.66$\times$} & \multirow{2}{*}{55.43} & \multirow{2}{*}{64.63} \\
\small{\textit{(voxel=0.1)}} & & & & \\
CenterPoint-Ada3D  & \multirow{2}{*}{0.31$\times$} & \multirow{2}{*}{0.25$\times$} & \multirow{2}{*}{58.62} & \multirow{2}{*}{65.68} \\
\small{\textit{(voxel=0.075)}} & & & & \\
\bottomrule
\end{tabular}}
\end{table}

\section{Implementation of sparse convolution}
\label{sec:sparse_conv}
We adopt the ``gather-scatter'' sparse convolution scheme in spconv~\cite{sparseconvnet}, torchsparse~\cite{torchsparse} and minkowskinet~\cite{minkunet} for the 3D and Ada3D optimized 2D sparse features. The sparse convolution is applied on sparse input features to aggregate the nonzero features in the local neighborhood. A mapping that maps the input coordinate to corresponding output neighboring coordinates is constructed for merging the orderless voxel features into matrix form for aggregation.

Given the voxelized input point cloud, which is an unordered set of voxels paired with features $\{(p_j,x_j)\}$, where $x_j \in \mathbb{R}^C$  is a C-dimensional feature vector. It pairs with the point $p_j \in \mathbb{Z}^{D}$ in $D$-dimension space. For a sparse convolution layer with kernel size $K$ and stride $s$, let $W \in R^{K^D \times C_{in} \times C_{out}}$ be its weight and $\Delta^{D}(K)$ be the kernel offset (e,g, $\Delta^3(3) = \{-1,0,1\}^3$), $K^D$ represents the local neighbor size (e.g., when $K=3$, $D=3$, $K^D=27$, which describes a $3\times3\times3$ 3D neighborhood). Using $\delta$ to index $\Delta^D(K)$ neighbors, we denote the weight for certain neighbor as $W_{\delta} \in \mathbb{R}^{C_{in} \times C_{out}}$. The convolution process could be described as:
\begin{equation}
x_k^{out} = \sum_{\delta \in \Delta^D(K)} \sum_{j} \mathds{1} \left[ p_j = sq_k+\delta \right](x_j^{in}\times W_{\delta}),
\label{equ:sparse_conv}
\end{equation}
where $p_j \in P_{in}$, $q_k \in P_{out}$, $P_{in}, P_{out}$ represents the input and output coordinates with nonzero features, 
 and $\mathds{1}$ is the binary indicator function. In the ``sparse convolution'' process, each nonzero input coordinate is multiplied and accumulated with all nonzero neighborhoods, resulting in growing nonzero output coordinates. ($P_{in} \subset P_{out}$). Another form of sparse convolution keeps the identical input and output coordinates and the sparse pattern remains unchanged ($P_{in} = P_{out}$), which is called the ``submanifold sparse convolution''.

The input-output mapping $\mathcal{M}=\{(p_j,q_k,W_{\delta})\}$ is constructed to obtain neighbors for convolution in an orderless set of voxels. The mapping $\mathcal{M}$ generates output coordinates $P_{out}$ given the input coordinates and convolution type, then the ``map search'' process finds all nonzero features for aggregation. In order to effectively examine whether the input coordinate $q_{j}+\delta$ is nonzero, the input coordinates are recorded with a hash table. The keys are the input coordinates $p_j$ and values are the input index $j$. The hash function is a flattening the coordinates of each dimension into an integer. The sparse convolution iterates through the mapping $\mathcal{M}$ and gathers all input features associate with the same weight matrix to generate a contiguous matrix. Then, the matrix-matrix multiplication is applied. Finally, the calculated results are scattered and accumulated to the corresponding output coordinates. 

Ada3D preserves sparsity for the 2D BEV features, allowing the utilization of the above mentioned sparse convolution technique to accelerate the 2D backbone. Additionally, Ada3D reduces the input coordinates size of 3D features, which subsequently reduces the size of the input-output mapping, and saves computation and memory for 3D sparse convolution. 


\section{Detailed hardware profiling results}
\label{sec:more_hardware}

We present detailed statistics of hardware profiling corresponding to the bar plots in hardware experiments section in the tables below.
From the hardware profiling analysis, we summarize the following findings about how to effectively accelerate voxel-based 3D detectors.


\textbf{Discrepancy between software estimated metrics and hardware tested metrics.} We calculate the computaional and memory cost of the model with commonly used FLOPs and memory. Following prior research~\cite{hardware_survey}, we sum all activation sizes as the memory cost of the model, as it accounts for the amount of external memory access, which corresponds to the energy cost. 
Specifically, when calculating FLOPs for 3D sparse convolution, we only count the nonzero neighborhood voxels' computation. However, there exists discrepancy between the software estimated metrics (e.g., FLOPs and memory) and hardware tested metrics (e.g., latency and peak memory). Firstly, as seen in Table.~\ref{tab:end2end_peak_memory}, the peak memory of ``Ada3D-B'' model on KITTI is compressed by 2.22$\times$ while the memory size is reduced by 4.93$\times$. We attribute this discrepancy to the input-output mapping $\mathcal{M}$ discussed in Sec.~\ref{sec:sparse_conv}, which takes up a large size and not taken into account when estimating memory. Secondly, as shown in Table.~\ref{tab:end2end_latency}, the Ada3D optimized model could achieve $4\sim8\times$ FLOPs reduction. However, the latency speedup is only $1.2\times \sim 1.6\times$. This discrepancy arises because the sparse convolution on GPU has the overhead of gather-scatter process, and the acceleration does not scale linearly with the drop rate.


\textbf{Both the 3D and 2D backbone requires acceleration.} The voxel-based 3D detectors consist of the 3D backbone that employs 3D sparse convolutions, and the 2D backbone that uses dense, normal convolutions. While The 3D sparse convolution has a small number of FLOPs and memory size (i.e., 10$\times$ less compared to the 2D backbone in centerpoint) but runs slowly on GPU. As shown in Table.~\ref{tab:end2end_latency}, the 3D backbone has approximately the same latency as the 2D backbone. PointAcc~\cite{pointacc} also points out this phenonmenon by comparing the 3D MinkowskiNet~\cite{minkunet} and 2D SqueezeSeg~\cite{squeezeseg} network, the former has 7$\times$ less FLOPs but runs 1.3$\times$ slower. Therefore, both the 3D and 2D backbone needs to be optimized. Ada3D optimizes the efficiency of both the 3D and 2D backbone while most prior research solely focus on the 3D~~\cite{sps-conv,not_all_neighbor} or 2D part~\cite{real_time_3d_det}. 


\textbf{Acceleration under different sparse rates.} As mentioned earlier in this section, while Ada3D achieves 5$\times$ data reduction for the 2d backbone, it only results in 1.3$\times$ latency improvement. In contrast, as shown in Table.~\ref{tab:3d_op} and Table.~\ref{tab:keep_rate}, when 3D voxel data is compressed by 1.5$\times$, the latency also improves by 1.5$\times$. The acceleration of sparse convolution on GPU does not follow a linear relationship with the sparse rate. As seen in Table.~\ref{tab:2d_op}, a sparse rate of $\leq 10\%$ is required to achieve a 1.5$\times$ latency improvement. The 3D voxels have a relatively high sparse rate (1E-4$\sim$1E-1), and further increasing the sparse rate could linearly improve the efficiency. Higher improvements in latency can be achieved by exploring higher sparsity or designing domain-specific hardware architectures.

\begin{table}[h]
\centering
\caption{\textbf{The density and keep ratio for 3D layers of ``Ada3D-B'' on KITTI dataset.} The ``Compress'' is the reciprocal of keep rate. }
\resizebox{1.0\linewidth}{!}{
\begin{tabular}{ccccc}
\toprule
\multirow{2}{*}{Layer} &  \multicolumn{2}{c}{Density} & \multirow{2}{*}{Keep Rate} & \multirow{2}{*}{Compress} \\
\cmidrule(lr){2-3}
 & Pre & Post & &  \\
\midrule 
3d\_conv\_1 & 0.0007 & 0.0005 & 71.43\% & 1.4000$\times$ \\
3d\_conv\_2 & 0.0098 & 0.0077 & 78.57\% & 1.2727$\times$ \\
3d\_conv\_3 & 0.0534 & 0.0305 & 57.12\% & 1.7508$\times$ \\
3d\_conv\_4 & 0.2198 & 0.1407 & 64.01\% & 1.5621$\times$ \\
3d\_conv\_5 & 0.2198 & 0.1407 & 64.01\% & 1.5621$\times$ \\
\midrule
2d\_conv\_1 & 1.0000 & 0.0883 & 8.83\% & 11.3250$\times$ \\
2d\_conv\_2 & 1.0000 & 0.1336 & 13.36\% & 7.4850$\times$ \\
2d\_conv\_3 & 1.0000 & 0.1045 & 10.45\% & 9.5694$\times$ \\
2d\_conv\_4 & 1.0000 & 0.1416 & 14.16\% & 7.0621$\times$ \\
2d\_conv\_5 & 1.0000 & 0.1777 & 17.77\% & 5.6275$\times$ \\
2d\_deconv\_1 & 1.0000 & 0.2116 & 21.16\% & 4.7256$\times$ \\
\bottomrule
\end{tabular}}
\label{tab:keep_rate}
\end{table}


\begin{table*}[htb]
\centering
\caption{\textbf{The latency and peak memory for 3D layers of ``Ada3D-B'' on KITTI dataset.} All statistics are tested on one RTX3090 GPU. The latency is measured in milliseconds (ms). The memory is measured in megabytes (MB).}
\resizebox{1.0\linewidth}{!}{
\begin{tabular}{ccccccccccccc}
\toprule
\multirow{2}{*}{Layer} &  \multicolumn{3}{c}{Peak Mem.(\textit{torchsparse})} & \multicolumn{3}{c}{Peak Mem.(\textit{spconv})} & \multicolumn{3}{c}{Latency (\textit{torchsparse})} & \multicolumn{3}{c}{Latency (\textit{spconv})} \\
\cmidrule(lr){2-4} \cmidrule(lr){5-7} \cmidrule(lr){8-10} \cmidrule(lr){11-13}
 & Pre & Post & Compress & Pre & Post & Compress & Pre & Post & Speedup & Pre & Post & Speedup \\
\midrule 
3d\_conv\_1 & 24.2 & 17.8 & 1.3560$\times$ & 21.5 & 16.5 & 1.3030$\times$ & 1.0201 & 0.9631 & 1.0592$\times$ & 0.8346 & 0.8269 & 1.0093$\times$ \\
3d\_conv\_2 & 63.4 & 48.6 & 1.3045$\times$ & 54.2 & 42.5 & 1.2753$\times$ & 1.0873 & 1.0540 & 1.0316$\times$ & 0.9846 & 0.8942 & 1.1011$\times$  \\
3d\_conv\_3 & 67.9 & 37.9 & 1.7916$\times$ & 59.4 & 34.6 & 1.7168$\times$ & 1.4157 & 1.0975 & 1.2899$\times$ & 1.6351 & 0.9301 & 1.7582$\times$ \\
3d\_conv\_4 & 52.9 & 34.1 & 1.5513$\times$ & 49.1 & 34.4 & 1.4273 $\times$ & 1.6740 & 1.0999 & 1.5220$\times$ & 1.9999 & 1.2674 & 1.5779$\times$ \\
3d\_conv\_5 & 52.9 & 34.1 & 1.5513$\times$ & 52.5 & 38.1 & 1.3780$\times$ & 1.6764 & 1.0738 & 1.5612$\times$ & 2.0017 & 1.2631 & 1.5848$\times$  \\
\bottomrule
\end{tabular}}
\label{tab:3d_op}
\end{table*}

\begin{table*}[htb]
\centering
\caption{\textbf{The latency and peak memory for 2D layers of ``Ada3D-B'' on KITTI dataset.} All statistics are tested on one RTX3090 GPU. The latency is measured in milliseconds (ms). The memory is measured in megabytes (MB).}
\begin{tabular}{ccccccc}
\toprule
\multirow{2}{*}{Layer} &  \multicolumn{3}{c}{Peak Mem.} & \multicolumn{3}{c}{Latency} \\
\cmidrule(lr){2-4} \cmidrule(lr){5-7}
 & Pre (\textit{pytorch}) & Post (\textit{spconv}) & Compress & Pre (\textit{pytorch}) & Post (\textit{spconv}) & Speedup  \\
\midrule 
2d\_conv\_1 & 207.1 & 24.4 & 8.4877$\times$ & 2.4975 & 1.0764 & 2.3202$\times$ \\
2d\_conv\_2 & 138.6 & 25.4 & 5.4567$\times$ & 1.3270 & 0.9627 & 1.3784$\times$ \\
2d\_conv\_3 & 138.6 & 21.3 & 6.5070$\times$ & 1.3202 & 0.9098 & 1.4511$\times$ \\
2d\_conv\_4 & 138.6 & 27.4 & 5.0584$\times$ & 1.3396 & 0.9960 & 1.3445$\times$ \\
2d\_conv\_5 & 138.6 & 33.2 & 4.1747$\times$ & 1.3329 & 1.2110 & 1.1006$\times$ \\
2d\_conv\_6 & 138.6 & 38.6 & 3.5906$\times$ & 1.3311 & 1.2617 & 1.0550$\times$ \\
2d\_deconv\_1 & 103.6 & 55.5 & 1.8667$\times$ & 0.5517 & 0.2815 & 1.9600$\times$ \\
\bottomrule
\end{tabular}
\label{tab:2d_op}
\end{table*}

\begin{table*}[htb]
\centering
\caption{\textbf{The end-to-end latency for Ada3D optimized models on KITTI and nuScenes dataset.} All statistics are tested on one RTX3090 GPU. The latency is measured in milliseconds (ms).}
\begin{tabular}{cccccccccccccc}
\toprule
\multirow{2}{*}{Model} & \multirow{2}{*}{Dataset} &  \multicolumn{6}{c}{Latency} \\
\cmidrule(lr){3-8} \cmidrule(lr){9-12}
 &  & 2D & \small{\textit{Opt. (2D)}} & 3D & \small{\textit{Opt. (3D)}} & Overall & \small{\textit{Opt. (Overall)}} \\
\midrule 
Centerpoint & KITTI & 9.7 & - & 26.2 & - & 35.9 & - \\
Centerpoint (Ada3D-B) & KITTI & 6.7 & 1.45$\times$ & 19.7 & 1.33$\times$ & 26.4 & 1.36$\times$ \\
Centerpoint (Ada3D-C) & KITTI & 5.8 & 1.68$\times$ & 16.9 & 1.55$\times$ & 22.6 & 1.59$\times$ \\
\midrule 
Centerpoint & nuScenes & 11.3 & - & 22.8 & - & 34.1 & - \\
Centerpoint (Ada3D) & nuScenes & 10.9 & 1.03$\times$ & 17.6 & 1.30$\times$ & 28.2 & 1.21$\times$ \\
\bottomrule
\end{tabular}
\label{tab:end2end_latency}
\end{table*}

\begin{table*}[htb]
\centering
\caption{\textbf{The peak memory for Ada3D optimized models on KITTI and nuScenes dataset.} All statistics are tested on one RTX3090 GPU. The latency is measured in milliseconds (ms).}
\begin{tabular}{cccccccccccccc}
\toprule
\multirow{2}{*}{Model} & \multirow{2}{*}{Dataset} &  \multicolumn{6}{c}{Peak Mem.} \\
\cmidrule(lr){3-8} \cmidrule(lr){9-12}
 &  & 2D & \small{\textit{Opt. (2D)}} & 3D & \small{\textit{Opt. (3D)}} & Overall & \small{\textit{Opt. (Overall)}} \\
\midrule 
Centerpoint & KITTI & 1003.7 & - & 1039.7 & - & 2042.8 & - \\
Centerpoint (Ada3D-B) & KITTI & 225.8 & 4.45$\times$ & 694.4 & 1.50$\times$ & 920.2 & 2.22$\times$ \\
Centerpoint (Ada3D-C) & KITTI & 149.9 & 6.70$\times$ & 540.2 & 1.92$\times$ & 690.1 & 2.96$\times$ \\
\midrule 
Centerpoint & nuScenes & 871.0 & - & 1051.2 & - & 1922.2 & - \\
Centerpoint (Ada3D) & nuScenes & 450.0 & 1.94$\times$ & 745.0 & 1.41$\times$ & 1195.0 & 1.61$\times$ \\
\bottomrule
\end{tabular}
\label{tab:end2end_peak_memory}
\end{table*}

\section{Limitations and Future work}

\textbf{Limitations} As discussed in Sec.~\ref{sec:more_hardware}, 
the 2D BEV backbone exhibits only moderate latency improvement at relatively low sparse rates (e.g., 30\%$\sim$50\%). Further exploration of higher sparsity and hardware designed tailored for utilizing the existing sparsity is necessary. Additionally, the Ada3d optimized model shows moderate performance decay with plain finetuning for recovery. To further enhance its performance, more advanced tunig techniques such as distillation could be employed.

We summarize the future directions as follows: 

\textbf{Further improving Ada3D's performance and efficiency.} Further increasing the sparse rate and adopting more advanced performance recovery techniques is worth exploring to enhance the effectiveness of Ada3D.

\textbf{Adapting Ada3D for more 3D detectors.} We apply Ada3D on the commonly used centerpoint model. The Ada3D could be easily embedded into other voxel-based 3D detectors (e.g., pointpillar, PV-RCNN). 

\textbf{Adapting Ada3D for more 3D perception tasks.} Ada3D exploits the spatial redundancy of the 3D point cloud data, which is also applicable in other scenarios, such as 3D Segmentation and Tracking. We primarily evaluate Ada3D on the 3D object detection task, by redesigning the predictor's ground-truth representation, Ada3D could easily adapt to other 3D perception tasks.

\end{document}